\pdfoutput=1

\documentclass[11pt]{article}

\usepackage[]{acl}

\usepackage{times}

\usepackage{booktabs}
\usepackage{graphicx}
\usepackage{tabu}
\usepackage{xcolor}
\usepackage{placeins}
\usepackage{times}
\usepackage{latexsym}
\usepackage{subcaption}
\usepackage{color,soul}
\usepackage{graphicx}
\usepackage{latexsym}
\usepackage{inconsolata}
\usepackage{tabularx}
\usepackage{xcolor}
\usepackage{multirow}
\usepackage{multicol}
\usepackage{graphicx}
\usepackage{floatrow}
\usepackage{makecell}
\usepackage{xargs} 
\usepackage{amsthm}
\usepackage{caption}
\usepackage{soul}
\usepackage{color,soul}
\usepackage{mathtools}
\usepackage{array}

\usepackage{amsmath}  
\usepackage{amsfonts} 

\usepackage[T1]{fontenc}

\usepackage[utf8]{inputenc}

\usepackage{microtype}

\usepackage{inconsolata}

\usepackage{inconsolata}
\usepackage{tabularx}
\usepackage{xcolor}
\usepackage{multirow}
\usepackage{multicol}
\usepackage{graphicx}
\usepackage{floatrow}
\usepackage{makecell}
\usepackage{xargs} 
\usepackage{amsthm}
\usepackage{caption}
\usepackage{soul}
\usepackage{color,soul}
\usepackage{mathtools}
\usepackage{array}
\usepackage{arydshln}
\definecolor{pink1}{HTML}{efedf5}
\definecolor{pink2}{HTML}{bcbddc}
\definecolor{pink3}{HTML}{756bb1}

\definecolor{blue1}{HTML}{deebf7}
\definecolor{blue2}{HTML}{9ecae1}
\definecolor{blue3}{HTML}{3182bd}
\usepackage{nicematrix}

\usepackage{booktabs}
\usepackage{tabu}
\usepackage{xcolor}
\usepackage{graphicx}
\title{Finding a Needle in the Adversarial Haystack: A Targeted Paraphrasing Approach For Uncovering Edge Cases with Minimal Distribution Distortion\\
\large
\textbf{{\color{red} This paper contains real-world cases which are offensive/hateful in nature.}}}

\newcommand{\aspace}{\hspace{1em}}

\author{
    Aly M. Kassem  \aspace 
    Sherif Saad  \aspace\\
     School of Computer Science, University of Windsor \aspace\\
     \texttt{\{kassem6,sherif.saad\}@uwindsor.ca}\\}

\begin{document}
\maketitle

\begin{abstract}
Adversarial attacks against Language models (LMs) are a significant concern. In particular, adversarial samples exploit the model's sensitivity to small input changes. While these changes appear insignificant on the semantics of the input sample, they result in significant decay in model performance. In this paper, we propose Targeted Paraphrasing via RL (TPRL), an approach to automatically learn a policy to generate challenging samples that improve the model’s performance. TPRL leverages FLAN-T5, a language model, as a generator and employs a self-learned policy using a proximal policy optimization to generate the adversarial examples automatically. TPRL's reward is based on the confusion induced in the classifier, preserving the original text meaning through a Mutual Implication score. We demonstrate \& evaluate TPRL's effectiveness in discovering natural adversarial attacks and improving model performance through extensive experiments on four diverse NLP classification tasks via Automatic \& Human evaluation. TPRL outperforms strong baselines, exhibits generalizability across classifiers and datasets, and combines the strengths of language modeling and reinforcement learning to generate diverse and influential adversarial examples.

\end{abstract}

\section{Introduction}

LMs have made impressive advancements in classification, question-answering, and machine translation. However, they are susceptible to adversarial attacks, which exploit their vulnerability to small input changes \cite{jia2017adversarial, jin2020bert, alzantot2018generating, wallace2019universal, jia2019certified, cheng2019robust}. These attacks introduce variations not encountered during training. Two approaches to address these vulnerabilities are data augmentation-based techniques \cite{liu_data_2020, wang2015s, kobayashi2018contextual, yu2018qanet} and adversarial training-based approaches \cite{zhu2019freelb, yoo2021towards, xu_lexicalat_2019}. Expanding training data using pre-designed samples generated by data augmentation methods can assist classifier training. However, generated samples may lack an adversarial nature \cite{altinisik2022impact}, leading to confusion and inaccurate classification. Adversarial training-based approaches \cite{xu_lexicalat_2019, le-etal-2022-semi, deng_valcat_2022, iyyer_adversarial_2018, alzantot2018generating} address this limitation by generating challenging examples. This improves the model's ability to handle difficult subsets of data, enhancing robustness and performance.

\begin{figure}[]
\centering
\includegraphics[width=0.75\textwidth]{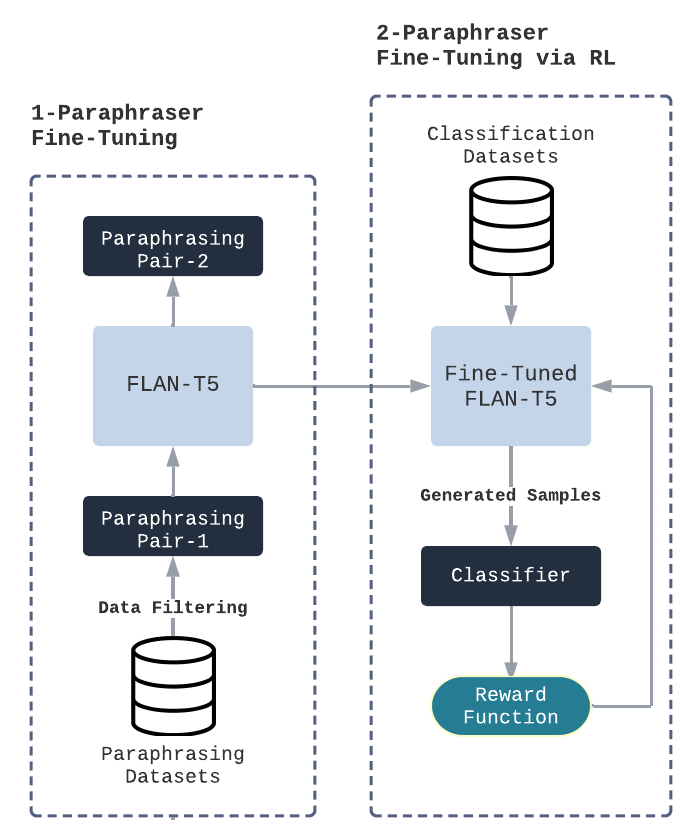}

\caption{Components of our framework \textbf{TPRL} for Natural Adversarial Generation. (1) Employing Data filtering and then paraphrasing fine-tuning. (2) Targetted paraphrasing through employing RL on classification Datasets.}
\label{fig:pipeline}
\end{figure}

Generating diverse and semantically meaningful adversarial examples is challenging due to limited operations like word addition, deletion, or substitution. This lack of diversity in word-level generation often results in generated sentences with identical vectors to the original, offering little insight into the model's behavior.

Recent studies \cite{le-etal-2022-semi, zhao_generating_2018} show that character manipulation or word swap methods can produce irrelevant and incoherent samples, altering the original text's meaning. These methods are unsuitable for real-world applications and can harm the model. Attacks lacking semantic significance expose the model's blind spots, are easily detectable and removable, and do not represent real-world examples encountered during deployment.

To address this, researchers have explored sentence-level generation methods. GAN-based approaches \cite{zhao_generating_2018, wang_cat-gen_2020} show promising results in generating diverse examples but may result in irrelevant and obtaining labeled data is challenging. Another approach employs a paraphrasing technique using LMs. However, this method lacks targeted fine-tuning to break the classifier and has limited available styles.
In this paper, we introduce a novel approach called TPRL that leverages FLAN-T5 for targeted paraphrasing, enhancing classifier performance through Reinforcement Learning. We employ a two-step process: first, we train a diverse paraphrasing model using FLAN-T5 and then fine-tune it with RL to preserve text meaning while generating targeted adversarial examples. We evaluate TPRL on four classification tasks, both automatically and with human evaluation, demonstrating its ability to improve classifier robustness and effectiveness. Our findings can be summarized as follows:
\begin{itemize}
    \item Utilizing the generated examples for adversarial training improves classifier performance on original and adversarial test sets.
    \item  Our work demonstrates that the learned policy for one classifier is universal and can be generalized to unseen classifiers in the same dataset.
    \item Experiments show that TPRL outperforms strong baselines and improves results on various models and datasets. 

\end{itemize}

\section{Background}
This section briefly introduces and formalizes textual adversarial attacks for text classification and employs RL in language models for generating adversarial attacks and other tasks.
\subsection{Textual Adversarial Attacks}
Generating adversarial attacks against NLP models is more challenging than for vision models \cite{qiu2022adversarial}. NLP models rely on discrete word representations, where even slight adjustments can drastically change the meaning or validity of a phrase. Unlike images, NLP models require a deeper understanding of context and language structure, making successful attacks difficult.

\textbf{Attacks Properties.} For a victim classification model, denoted as $F_\theta$, tested on dataset $D_t$ with samples $(x_t, y_t)$, "an adversarial attacker aims to perturb $x_t$ to maintain semantic similarity to humans but destroy its meaning when classified by the model." This generates an adversarial example, $x^{'}_{t}$, that the model misclassifies. 

\subsection{RL In Language Models}
Dynamically generating adversarial attacks at the sentence level is more compatible with the reinforcement learning (RL) paradigm. In the realm of NLP, RL has gained prominence for addressing undesirable behavior, including toxicity, social biases, offensive speech, and data memorization\cite{paulus2017deep, rennie2017self, wu2016google, kassem-etal-2023-preserving}. This is accomplished by using Proximal Policy Optimization (PPO) \cite{schulman2017proximal} to optimize a Language Model (LLM) based on a reward model. Despite RL's success in mitigating undesirable behavior in LMs, its potential for generating adversarial attacks remains largely unexplored. This paper presents the first investigation of using RL with a language model for generating natural adversarial attacks.
 \section{Collecting Labeled Paraphrasing Pairs}
This section will show the selected datasets for training the paraphrase model. Afterward, we will outline a systematic procedure for filtering those datasets to maximize the paraphrase pairs' diversity, similarity, and relevance.

\subsection{Paraphrasing Datasets}
In our initial stage, we gathered seven diverse paraphrase datasets, most undergoing thorough human judgment annotation to ensure high-quality paraphrasing examples. This comprehensive paraphrasing corpus comprises data from the APT dataset \cite{nighojkar2021improving}, Microsoft Research Paraphrase Corpus (MSRP) \cite{dolan-brockett-2005-automatically}, PARANMT-50M corpus \cite{wieting-gimpel-2018-paranmt}, TwitterPPDB \cite{lan2017continuously}, PIT-2015 \cite{xu-etal-2015-semeval}, PARADE \cite{he2020parade}, and QQP \cite{iyer2017first}. We utilized a filtered version of the PARANMT-50M corpus as recommended by \cite{krishna2020reformulating}. We retained sentence pairs with 4, 5, and 6 similarity labels from TwitterPPDB and sentences with semantic similarity labels of 5 and 4 from PIT-2015. For QQP, we selected samples labeled as duplicates. The merged dataset, totaling 560,550 samples, underwent filtering to ensure high-quality similarity and diversity for the subsequent stage.

\subsection{Improving Diversity \& Relevance Via Data Filtering}
In the second stage, we choose training data for the paraphrase model, following \cite{krishna2020reformulating}. Despite the availability of human annotations, it is still possible for noise and irrelevant samples to exist in the dataset. We employ aggressive filtering with four filters to address noise and irrelevant samples in the dataset. Firstly, we remove sentence pairs with over 50\% unigram overlap, ensuring lexical diversity computed using SQUAD evaluation scripts based on the F1 score \cite{rajpurkar-etal-2016-squad}. Secondly, we discard pairs with less than 50\% reordering of shared words, promoting syntactic diversity measured by Kendall's tau \cite{kendall1938new}. Thirdly, we eliminate pairs with less than 50\% semantic similarity, measured by cosine similarity using the "all-MiniLM-L12-v2" model \cite{wang2020minilm, reimers2019sentence}. Finally, we remove sentences with over 70\% trigram overlap to improve diversity further. After applying these filters, the refined dataset contains 96,073 samples, split into training (76,857 samples), validation (9,608 samples), and testing (9,608 samples) sets. These filters ensure a diverse and representative sample for effective training and evaluation.


\section{Targeted-Paraphrasing Via RL}
\autoref{fig:pipeline} illustrates the framework's structure. After filtering data for diverse and relevant paraphrase pairs, we proceed with the initial fine-tuning of the model. Subsequently, we utilize proximal policy optimization, a reinforcement learning technique, to fine-tune the model further. This approach generates paraphrases that exploit the classifier's weaknesses, resulting in complex and effective adversarial samples.

\subsection{Paraphraser Model}
We enhance the FLAN-5-large model by fine-tuning the filtered data for nine epochs. Employing the BERT-Score metric \cite{zhang2019bertscore}, it achieves an F1-score of 75.925\%, enhancing fluency, diversity, relevance, and paraphrasing ability. The LM also addresses task-irrelevant generation issues. Training on relevant pairs maximizes task-specific outputs. Utilizing the LM boosts paraphrasing capability and introduces new information about entities or objects in the input text, improving generation quality \cite{petroni2019language}. This optimized paraphrase is then used to create adversarial training samples.

\subsection{Fine-Tuning Paraphraser Via RL}
After the initial fine-tuning, the paraphraser model can generate various relevant and fluent paraphrases. To enhance its performance, our approach includes a guiding component via the reward function. This involves further fine-tuning the model using reinforcement learning (PPO), enabling it to produce targeted adversarial examples that confuse the classifier.

\subsubsection{Reward Function}
Given tokens \(x_{<t} = \{x_0, x_1, \ldots, x_{t-1}\}\) and accumulated hidden states \(h_{\theta<t}\) before time step \(t\). An auto-regressive language model (LM) is trained to maximize the probability of the next step token \(\hat{x}_t\). LM as a generator $G$ selects the token that has the highest probability \(x_t\) as the \(t\)-th step decoding output:
\begin{equation}
 x_t \sim \text{argmax}_{\hat{x}_t} p(\hat{x}_t | x_{<t}) = \text{G}(x_{<t}, h_{\theta<t}) 
\end{equation}

In the reinforcement learning framework, we define the state at step \textit{t} as all the sequences generated before t \( s_t = x_{<t} \), and the action at step t as the t-th output token (\( a_t = x_t \)). The policy \( \pi_\theta\) represents the probability of selecting token $x_t$ (action \(a_t\)) given the preceding state 
\(s_t = x_{<t}\). This probability is derived from the softmax output of the hidden states \( \pi_\theta(a_t|s_t) = \text{softmax}(h_{\theta<t}) \), and this interpretation extends to the conditional case as well.
The single-step reward for token \(x^c_t\) at step \(t\) can defined as follows:
\begin{equation}\label{equation-2}
    R(x^c{_t}) = E_t \left[ \frac{\pi_{\theta_c}(a_t|s_t)}{\pi_\theta(a_t|s_t)} r(x^c{_t}) \right] - \beta \text{KL}(\theta||\theta_c)
\end{equation}
Where \(r(x^c{_t})\) is the objective composed of the weighted sum of confusion and Mutual Implication. The KL penalty is applied per token using a reference model, which is the original model that does not receive the signal reward to prevent significant deviations. 

\textbf{Confusion constraint with classifier.} 
To generate confusing samples that challenge the model, we randomly select pairs $(x, y)$ from the dataset $D$ and input them into the generator $G$. This process yields a novel pair $(\hat{x}, y)$ through generation. Subsequently, we fed these generated instances $(\hat{x}, y)$ into the classifier $C$ to obtain the likelihood of the true label $L(y|\hat{x}) = p_C(y|\hat{x}; \psi)$. Where $\psi$ represents the parameter of the classifier. We then incorporate $(1-L(y|\hat{x}))$ into our reward function. This term aims to incentivize the generator to acquire a policy that produces confusing samples capable of decreasing the classifier's confidence. It can be viewed as searching for confusing samples within the classifier's space.

\textbf{Similarity constraint with Mutual Implication.} 
Previous research in word and sentence-level attacks often used embeddings like word2vec \cite{mikolov2013efficient}, and GloVe \cite{pennington2014glove}, along with contextual embeddings such as BERTScore \cite{zhang2019bertscore} and BLEURT \cite{sellam2020bleurt}. However, these methods often fail to capture inferential semantics, leading to irrelevant sample generation when used as a reward function. For example, BERTScore assigns a high similarity score to sentences like "I bought an iPad" and "I bought a Laptop," even though they refer to different products.

To address this, we propose an alternative approach: using Mutual Implication \cite{nighojkar2021improving}, a similarity score that more accurately evaluates text generation by considering both relevance and semantic equivalence as the bi-directional relationship adds more constraints between the generated and original sentence. Let $A = a1,\ldots, an$ and $B = b1, \ldots, bm$ be two token sequences, and let $L_{NLI}$ be a language model trained on the Natural Language inference task (NLI) that produces three labels, Entailment, Contradiction, and Neutral. We estimate the Mutual Implication as follows:
\begin{equation}
    \text{MI}(A, B) = [L_{NLI}(A \Rightarrow B) + L_{NLI}(B \Rightarrow A) ] /2. 
\end{equation}

 Where $L_{NLI}$ is ALBERT XXLarge $v_2$ model \cite{lan2019albert} trained on various NLI datasets \cite{nie2019combining, bowman2015large, williams-etal-2018-broad, nie2019combining, nie-etal-2020-adversarial}.\newline
 
\textbf{KL Penalty.} The policy of the fine-tuned model may deviate significantly from the old policy (the model before fine-tuning), potentially leading to a less coherent and relevant generation. To address this issue, we introduce a KL divergence penalty term to quantify the dissimilarity between these two policies. This step helps ensure that our optimization process remains within a trustworthy region. The KL divergence, calculated for the policies, is expressed as:
\begin{equation}
KL(\theta || \theta_c) = \sum_{i \in [1, t]} \pi_\theta(a_i | s_i) \cdot \log \frac{\pi_\theta(a_i | s_i)}{\pi_{\theta_c}(a_i | s_i)}
\end{equation}
We deduct KL divergence with default value weight $\beta=0.2$ as a penalty term in the reward function (\autoref{equation-2}).

\textbf{Objective function.} 
We consolidate all constraints from previous sections into a unified objective. Our objective function blends the negative likelihood, which drives the generator to produce confusing samples, with the mutual implication, which ensures the similarity, naturalness of meaning, and semantic equivalence. This measures how closely the paraphrases convey the same meaning as the original text and vice versa. The final objective function is defined as:
\begin{equation} r(\hat{a_t}) = \upsilon \cdot (1 - p_C(y|\hat{x}; \psi)) + \alpha \cdot MI(x,\hat{x}) \end{equation}
Here, $\psi$ represents the classifier's parameter, and $\upsilon$ as well as $\alpha$ serve as weighting factors, both set to 0.5. This balance ensures that the model neither generates overly confusing examples at the expense of relevance nor produces irrelevant results at the cost of coherence.
\subsubsection{Generation Settings}
We applied several constraints to sentence generation, such as matching the generated sentence length to the original one to avoid irrelevant tokens. We use early stopping when enough complete candidate beams are available to prevent meaningless or off-topic tokens. These constraints enable the generator to introduce new, relevant information or replace existing tokens with a diverse range of novel and pertinent tokens. We also generate ten alternative samples and assess their adequacy and fluency using the Parrot Tool \cite{prithivida2021parrot}, selecting the most suitable one for reward computation to maximize fluency and relevance.

\subsubsection{Policy Optimization}
We use a proximal policy optimization approach with top-p sampling 0.95, known as Natural Language Policy Optimization \cite{ramamurthy2022reinforcement}(see \autoref{appendix A} for more details). Given the reward and the definitions described above, we update our policy at $t$-th step as:
\begin{equation}
\resizebox{1\hsize}{!}{$\theta_{\text{new}} = argmax_{\theta} \mathbb{E}\left[\min\left(r_t(\theta), \text{clip}\left(r_t(\theta), 1 - \epsilon, 1 + \epsilon\right)\right) A_t\right]$}
\end{equation}
where $r_t(\theta) = \frac{\pi_{\theta}(a_t|s_t)}{\pi_{\theta_{\text{old}}}(a_t|s_t)}$. The optimization objective is to find the new policy parameters that maximize expected rewards while keeping the policy update bounded within a certain range defined by the clipping parameter. This helps maintain stability during training. PPO also balances the trade-off between exploration and exploitation by encouraging actions that have higher estimated advantages while avoiding drastic policy changes that could disrupt learning. Further details on PPO \& NLPO methods can be found in \autoref{appendix A}. The model was trained for thirty epochs with a batch size of 32 and optimized using the Lion optimizer \cite{chen2023symbolic}, using a learning rate of $4.9 \times 10^{-6}$; see \autoref{appendix hyper} for more details about the hyperparameters.

\subsection{Sampling Adversarial Texts} 
Once the paraphraser has been optimized vial RL, we can sample from the adversarial distribution to construct adversarial examples. We generate counterparts for each sample in the original training set to create adversarial samples. However, it is possible that some samples are irrelevant or not useful for training \cite{xu_lexicalat_2019}. Thus, we exclude those with a mutual implication score below 50\%, ensuring that only semantically equivalent samples are included. Finally, These samples are added to the training set, creating an updated dataset. Using the same random seed, we train a new classifier from scratch, employing adversarial training.

For the classifiers, we began by fine-tuning the classification models (\autoref{subsection:cls}) on the respective dataset (\autoref{subsection:Datasets}) and reported their performance. Each model achieved a different performance and made different errors. We then leveraged these classifiers within the RL feedback loop.




\section{Experiments}
We assess the effectiveness of targeted paraphrasing via RL adversarial attacks (TPRL) on four distinct classification tasks: sentiment analysis, news topic classification, hate speech detection, and offensive speech detection. To this end, we carefully select relevant datasets, outline our implementation details, establish baseline methods, and specify evaluation metrics. 

\subsection{Experimental Settings}

\subsubsection{Tasks \& Datasets.} \label{subsection:Datasets}

\textbf{Sentiment Analysis.} \textit{SST-2 \& SST-5} datasets for sentiment analysis in movie reviews from the Stanford Sentiment Treebank \cite{socher2013recursive}. SST-2 (N=6920)has binary sentiment labels (positive or negative), while SST-5 (N=8540) has more fine-grained sentiment labels (very positive, positive, neutral, negative, and very negative) with an average of 19 words per sample.

\textbf{New Topic Classification.} \textit{AG News} dataset \cite{zhang2015character} with a number of samples 120,000, categorizing news articles into four classes: World, Sports, Business, and Science/Technology, with an average of 38 words per sample.

\textbf{Offensive Speech Detection.} \textit{SemEval2019} Task 6 (OffensEval) dataset (N=11916) \cite{zampieri2019semeval} for offensive detection in tweets has binary classes: offensive and non-offensive tweets
with an average of 19 words per sample.

\textbf{Hate Speech Detection.} \textit{Hate speech} dataset (N=7071), a collection of sentences extracted from Stormfront, a white supremacist forum
\cite{de-gibert-etal-2018-hate}. Based on their content, sentences are categorized into two different classes, HATE and NoHate, with an average of 16 words per sample.

We eliminated all punctuation, mentions, hashtags, and URL links from the samples of all datasets. Furthermore, we employed lowercase for all samples. The maximum sequence length was implemented as the maximum length parameter in all BERT models. The training, validation, and test sets officially released by the creator of the datasets were utilized.


\begin{table}[!h]
\centering
\resizebox{7.4cm}{!}{%
\begin{tabular}{lcccccc}
\toprule
\multirow{1}{*}{\textbf{Classifier}} & \multicolumn{1}{c}{\textbf{SST2}} & \multicolumn{1}{c}{\textbf{SST5}} & \multicolumn{1}{c}{\textbf{AG’s News}} & \multicolumn{1}{c}{\textbf{HS\textsuperscript{$\ddagger$}}} & \multicolumn{1}{c}{\textbf{OFF\textsuperscript{$\ddagger$}}} \\

\toprule 

BERT$_{Base}$ & $90.88$ & $53.52$ & $93.97$ & $\star$ & $84.76$ \\
+\textit{SCPN} & $89.67$ & $51.71$ & $93.26$ & $\star$ & $83.60$ \\
+\textit{StyAdv} & $87.91$ & $52.35$ & $93.19$ & $\star$ & $81.86$ \\
+\textit{UNTP} & $\star$ & $51.90$ & $94.17$ & $\star$ & $\star$ \\
\textbf{+\textit{TPRL}} & $\bf 91.15$ & $52.39$ & $\bf 94.46$ & $\star$ & $\bf 85.11$ \\
\\

BERT$_{Large}$ & $91.43$ & $53.52$ & $94.18$ & $\star$ & $85.11$ \\
+\textit{SCPN} & $90.66$ & $53.61$ & $93.17$ & $\star$ & $72.09$ \\
+\textit{StyAdv} & $90.17$ & $23.07$ & $93.57$ & $\star$ & $72.09$ \\
+\textit{UNTP} & $\star$ & $54.84$ & $\bf 94.42$ & $\star$ & $\star$ \\
\textbf{+\textit{TPRL}} & $\bf 92.58$ & $\bf 54.93$ & $94.36$ &$\star$ & $\bf 85.58$ \\
\\

RoBERTa$_{Base}$ & $94.34$ & $54.79$ & $93.78$ & $91.90$ & $83.95$ \\
+\textit{SCPN} & $92.31$ & $53.89$ & $93.61$ & $91.30$ & $82.67$ \\
+\textit{StyAdv} & $91.81$ & $52.21$ & $93.32$ & $91.55$ & $82.32$ \\
+\textit{UNTP} & $\star$ & $\bf 56.19$ & $93.78$ & $91.90$ & $\star$ \\
\textbf{+\textit{TPRL}} & $94.00$ & $56.15$ & $\bf 93.93$ & $\bf 92.45$ & $\bf 85.00$ \\
\\

RoBERTa$_{Large}$ & $93.73$ & $58.30$ & $93.92$ & $92.45$ & $85.93$ \\
+\textit{SCPN} & $49.91$ & $23.07$ & $93.80$ & $92.30$ & $72.093$ \\
+\textit{StyAdv} & $49.91$ & $23.07$ & $93.73$ & $91.75$ & $72.09$ \\
+\textit{UNTP} & $\star$ & $\star$ & $93.86$ & $92.45$ & $\star$ \\
\textbf{+\textit{TPRL}} & $\bf 94.72$ & $\bf 58.95$ & $\bf 94.21$ & $92.05$ & $\bf 84.53$ \\
\\

DeBERTa-V3$_{Large}$ & $94.89$ & $58.46$ & $93.92$ & $89.50$ & $84.65$ \\
+\textit{SCPN} & $93.52$ & $\bf 59.23$ & $93.75$ & $89.50$ & $84.76$ \\
+\textit{StyAdv} & $93.41$ & $55.42$ & $\star$ & $\bf 92.95$ & $80.93$ \\
+\textit{UNTP} & $\star$ & $58.09$ & $93.85$ & $89.50$ & $\star$ \\
\textbf{+\textit{TPRL}} & $\bf 95.82$ & $58.77$ & $\bf 94.22$ & $92.30$ & $\bf 85.93$ \\
\bottomrule
\end{tabular}%
}
\caption{The classifier-dataset experiments for five classifiers. We show the accuracy results on the original test set before \& after we apply AT with TPRL \& the three baseline methods. \textsuperscript{$\ddagger$} Refer to Hate dataset, OFF to Offensive dataset. \textbf{*} Refer to not deployed experiments. The best comparable performances are bolded}
\label{table:comp_lfs}
\end{table}

\subsubsection{Victim Models.}\label{subsection:cls}
To assess the effectiveness of our approach across different models, we selected five popular pre-trained language models: \textit{BERT-base}, \textit{BERT-large} \cite{devlin2018bert}, \textit{RoBERTa-base}, RoBERTa-large \cite{liu2019roberta}, and DeBERTa-v3-large \cite{he2021debertav3}, which vary in architecture and size.
\newline

\textbf{Baseline Methods.}
TPRL is compared to SCPN \cite{iyyer_adversarial_2018} and StyleAdv \cite{qi_mind_2021}, other sentence-level adversarial attack techniques. SCPN uses a seq2seq Bi-directional LSTM \cite{hochreiter1997long} with the PARANMT-50M corpus and syntactic templates. StyleAdv utilizes the STRAP model \cite{krishna2020reformulating} for style transfer, incorporating five distinct styles. Also, we considered an untargeted paraphrasing model (UNTP) without the guiding component. See \autoref{appendix B} for implementation details.


\subsection{Evaluation Metrics}
We thoroughly evaluated TPRL's effectiveness in five key areas, ensuring the following:

\textbf{(1) Improving Performance:} We evaluated TPRL's impact on the accuracy of the original test set, alone and in combination with other methods, to assess its overall performance enhancement. Emphasizing that we follow a model-performance perspective, our objective is to find the model's weaknesses and enhance them, not just attack or break the model. 

\textbf{(2) Fluency and Quality:} We assessed fluency using perplexity (PPL) from GPT-2-XL \cite{radford2019language} and a RoBERTa-large classifier trained on the CoLA corpus \cite{warstadt2019neural}, to overcome the limitations of perplexity in evaluating fluency as the model provides accurate grammatical acceptability judgments.

(3) \textbf{Semantic Similarity (SIM):} We assessed semantic similarity between input sentence and generated samples using the "all-MPNet-Base-v2" embedding-based SIM model \cite{song2020mpnet, reimers2019sentence}, known for its performance on semantic textual similarity (STS) benchmark \cite{muennighoff2022mteb}. We also used the mutual implication (MI) metric to capture inferential semantics comprehensively, addressing the limitations of STS.

(4) \textbf{Validity Via Human Evaluation:} We conducted human evaluations to determine the percentage of samples that produced adversarial examples without altering the original label. 

(5) \textbf{Validity Via GPT-3.5:} To overcome human evaluation cost, we employed GPT-3.5, which has demonstrated comparable or superior performance to crowd-workers in-text annotation tasks \cite{gilardi2023chatgpt, tornberg2023chatgpt, chiang2023vicuna}. This assessment validated the credibility of TPRL-generated samples.

\begin{table}
\centering
\resizebox{7cm}{!}{%
\begin{tabular}{ccccc}
\hline
\multirow{2}{*}{\textbf{Classifier/Framework}} & 
\textbf{SCPN} &
\textbf{StyleAdv} &
\textbf{UNTP} &
\textbf{TPRL}  \\
& (\%) & (\%) & (\%) & (\%)\\
\hline
\textbf{BERT$_{Base}$} & 69.37 & 30.62 & 64.43  & 34.84\\
+\textbf{AT} &68.70 &  54.28 & 66.24 &42.28\\
\hline
\textbf{BERT$_{Large}$} & 70.86 & 29.97 & 64.89  &41.51\\
+\textbf{AT} & 68.00 & 51.73 & 64.35 &  44.36\\
\hline
\textbf{RoBERTa$_{Base}$} & 95.53 & 29.29 & 67.52  &37.27\\
+\textbf{AT} & 74.01 & 57.42 & 67.27 &  43.41\\
\hline
\textbf{RoBERTa$_{Large}$} & 82.49 & 28.02 & 89.78  &40.27\\
+\textbf{AT} & 75.55 & 52.28 & 90.19 &  49.22\\
\hline
\textbf{DeBERTa-v3$_{Large}$} & 74.77 & 36.79 & 63.38  &34.66\\
+\textbf{AT} & 75.01 & 45.33 & 64.83 &  47.82\\
\hline
\end{tabular}
}
\caption{\label{advs_res_tbl} The classifier-framework experiments for five classifiers. We show the accuracy results on the adversarial test set before \& after we apply AT with TPRL \& the three baseline methods.}
\end{table}

\subsection{Experimental Results \& Discussion}
We conducted comprehensive experiments to answer the following three overarching questions regarding TPRL:
\subsubsection{Does TPRL Enhance The Performance?}
\autoref{table:comp_lfs} shows the evaluation of TPRL’s accuracy against the three baseline methods and vanilla models on different benchmark datasets. Overall, TPRL consistently gained improved performance across the five classifiers and datasets. While UNTP outperformed SCPN and StyleAdv, it still lacked consistent effectiveness. In terms of the adversarial test set as shown in \autoref{table:rf}, SCPN did not enhance classifier performance, StyleAdv showed significant improvement but not consistently on the original test set, and UNTP had a small and sometimes non-existent improvement. However, TPRL achieved an $\sim$8\% improvement on the adversarial test set while maintaining a balance in the tradeoff between the performance on both the original and adversarial sets. Note that the missing values in \autoref{table:comp_lfs} because UNTP failed to meet the filtering criteria (<0.5 MI) for the SST2, Hate speech, and offensive speech datasets. Generator collapse\footnote{The language model-generator could not learn an effective policy to attack the classifier and produced such low reward values.
} occurred during training for BERT$_{Base}$ in the hate speech dataset. For a fair comparison, AG's News with DeBERTa-V3$_{Large}$ was excluded due to low accuracy in adversarial training.

\begin{figure}[]
\centering
\frame{\includegraphics[width=1.0\textwidth]{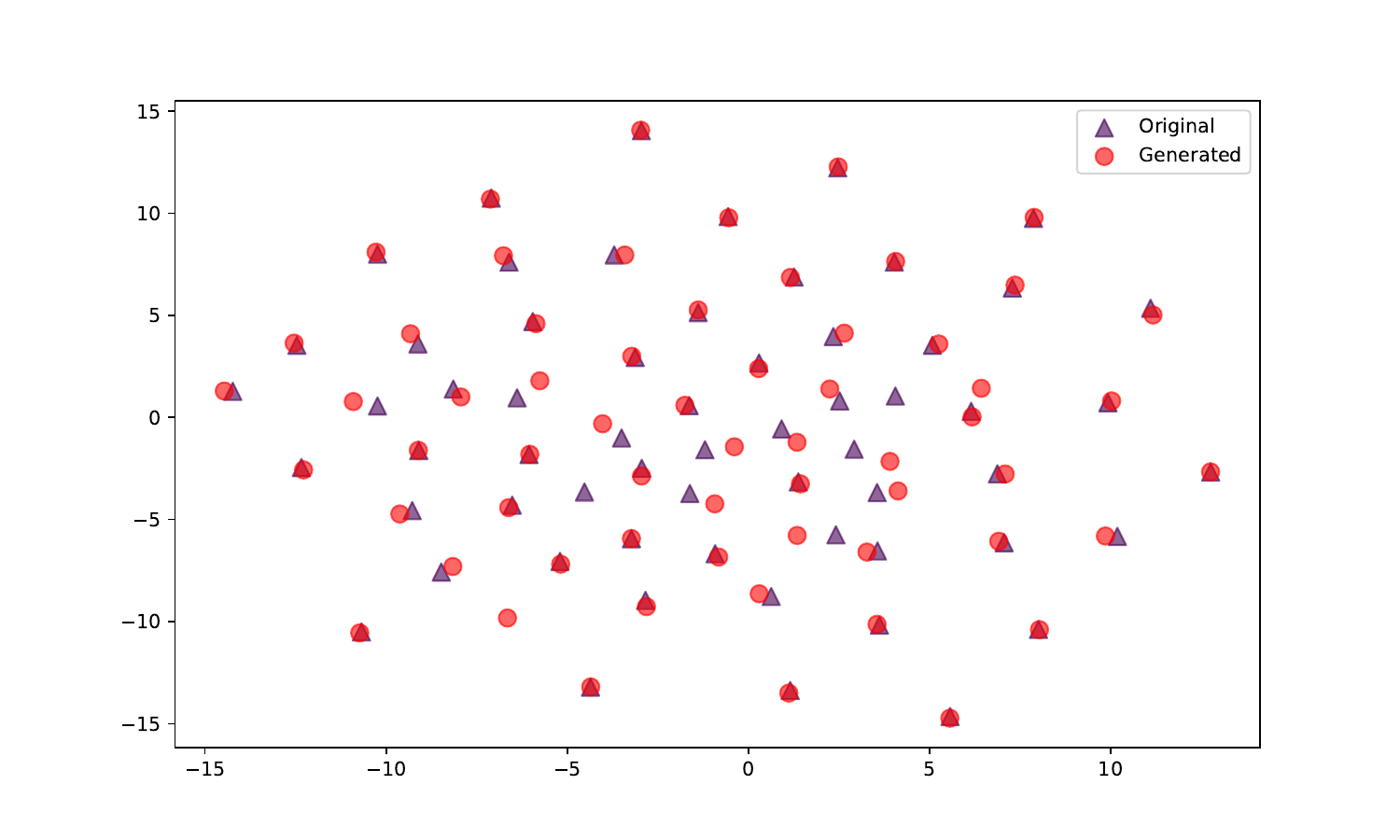}}

\caption{T-SNE visualization of the vectorized original and TPRL-adversarial sentences in the SST-2. The adversarial sentences (circles) mostly overlap with the original sentences (triangles), suggesting that generated sentences maintain the original class distribution.}
\label{fig:tsne}
\end{figure}

\subsubsection{Does TPRL Generate Relevant Samples?}
In an adversarial generation, maintaining topic and meaning while fooling the classifier is crucial since a sample can easily change the classifier's decision if the meaning changes. This concern is even more significant for sentence-level methods that create new sentences. We evaluated the generated samples using three approaches: 

\begin{table}[htbp]
\begin{center}
    \centering
    \captionsetup{justification=centering}
    \caption{Automatic evaluation results showing the average of generated adversarial training samples of the five datasets across selected classifiers \& baselines. The best comparable performances are bolded}
    \label{table:rf}
    \small
    \resizebox{7cm}{!}{%
        \centering
        \begin{tabular}{cccccc}
            \toprule
            \multirow{2}{*}{\textbf{Classifier}} & \multirow{2}{*}{\textbf{Framework}}  & $\bf PPL\downarrow$ & $\bf FL\uparrow$ & $\bf SIM\uparrow$ & $\bf MI\uparrow$\\
            & & (\%) & (\%) & (\%) & (\%)\\
            \midrule
            \multirow{4}{*}{BERT\textsubscript{Base}}
            & SCPN & 567.66  & 50.57 & 74.49 & 80.58\\
            & StyleAdv & 670.47 &57.65&74.62& 58.34\\
            & UNTP & 498.61 & $85.20$ & $\bf77.33$ & $\bf 92.41$ \\
            & TPRL & $\bf 368.41$ & $\bf 87.1$ & $73.56$ & $89.9$ \\
            \midrule
            \multirow{4}{*}{BERT\textsubscript{Large}}
            & SCPN & 565.61  & 50.71 & 74.67 & 80.94\\
            & StyleAdv & 863.50 &56.42& 74.14 & 57.40\\
            & UNTP & 373.52 & 85.97 & $\bf 77.64$ & $\bf 91.25$ \\
            & TPRL & $\bf 372.51$ & $\bf 86.84$ & 73.88 & 89.82 \\
            \midrule
            \multirow{4}{*}{RoBERTa\textsubscript{Base}}
            & SCPN & 563.46  & 50.31 & 74.43 & 80.28\\
            & StyleAdv & 708.28 &57.26&75.66 & 58.99\\
            & UNTP & $\bf 254.37$ & 83.31 & $\bf 78.32$ & 89.92 \\
            & TPRL & 492.52 & $\bf 85.58$ &71.22 & $\bf 89.99$ \\
            \midrule
            \multirow{4}{*}{RoBERTa\textsubscript{Large}}
            & SCPN& 555.02  & 50.43 & 74.43 & 80.19\\
            & StyleAdv & 579.80 &57.34&74.30 & 56.81\\
            & UNTP & $\bf 230.73$ & 83.32 & 78.44 & 88.49 \\
            & TPRL & 302.76 & $\bf 87.38$ & $\bf 70.84$ & $\bf 90.61$ \\
            \midrule
            \multirow{5}{*}{DeBERTa-v3\textsubscript{Large}}
            & SCPN & 560.70  & 50.29 & 74.28 & 79.91\\
            & StyleAdv & 845.76 & 57.69 & $\bf 74.92$ & 57.46\\
            & UNTP & $\bf 372.44$ & 73.21 & 68.86 & 73.14 \\
            & TPRL & 393.06 & $\bf 87.27$ & 73.90 & $\bf 89.67$ \\
            \bottomrule
        \end{tabular}%
    }
    \normalsize
\end{center}
\end{table}

\textit{(1) Automatic evaluation for similarity and fluency, including PPL for accurate fluency assessment.} \textit{(2) Human evaluation for validity.} \textit{(3) Visualization techniques for observing the geometric interpretation of samples.} \autoref{advs_res_tbl} shows the results, with TPRL and UNTP achieving the lowest PPL and superior generation quality. TPRL also outperforms other baselines in fluency, ranging from 86\% to 87\%, across classifiers and datasets, thanks to including the MI score in the reward function to encourage natural sentence generation.

\begin{table*}
\begin{center}
    \centering
\resizebox{15cm}{!}{%
\begin{tabular}{c|ccccc}
\hline

\multirow{2}{*}{\textbf{Policy/Classifier}}& 
\textbf{BERT$_{Base}$} & \textbf{BERT$_{Large}$} & \textbf{RoBERTa$_{Base}$} & \textbf{RoBERTa$_{Large}$} 
& \textbf{DeBERTa-v3$_{Large}$} \\
  & (\%) & (\%) & (\%) & (\%) & (\%)\\
\hline
None & 90.88 & 91.43 & 94.34 & 93.73 & 94.89
\\
Policy-BERT$_{Base}$ & 91.15 & 92.09& 93.35 & $\star$& 95.38
\\
Policy-BERT$_{Large}$ & 91.04 & 92.58 & \bf 94.45& 94.94& \bf 96.15
\\
Policy-RoBERTa$_{Base}$ & 90.06 & \bf 93.24 & 94.0& $\star$ & 95.60
\\
Policy-RoBERTa$_{Large}$ & \bf 91.87 & 92.36 & $\star$ & 94.72 & 95.60
\\
Policy-DeBERTa-v3$_{Large}$ & 90.93 & \bf 93.24 & 94.12 & \bf 95.49 & 95.82
\\
\hline
\end{tabular}
}
\caption{\label{unv_policy_table}
Accuracy results of different classifiers trained with the examples generated by various attacking policies on the SST-2 dataset. Showing the universal policy. The best comparable performance policy for the classifier is bolded}

\end{center}

\end{table*}

\textbf{Regarding The Relevance:} TPRL surpasses baseline methods in MI. For specific configurations, cosine similarity scores are low ($\sim$74\%), while MI scores are high ($\sim$89\%). This discrepancy arises because cosine similarity struggles to capture the inferential role accurately (further explanation given in \autoref{sec:abstudies}). See \autoref{appendix e} for dataset-specific results.\\
\textbf{For Human evaluation:} Following \cite{qi_mind_2021}, considering the cost, we conducted a validity evaluation on SST2. We randomly selected 100 adversarial samples for TPRL, SCPN, and StyleAdv (36, 33, and 31 samples, respectively). Each sample was evaluated by three annotators who determined if the sentiment matched the original example. The final decision was made by voting. The percentage of valid adversarial samples was TPRL 72\%, SCPN 51.5\%, and StyleAdv 32.2\%. TPRL achieved the highest validity, confirming minimal distortion to the original distribution. See \autoref{appendix:had} for details about human evaluation.

\textbf{For Validity Via GPT-3.5.} To evaluate the similarity of larger generated samples, we employed GPT-3.5. Randomly choosing 100 samples from each framework in the SST-2 dataset, we rated them on a scale of 1 to 5, where 1 indicated significant dissimilarity, and 5 denoted substantial similarity. TPRL exhibited superior performance to the three baselines, receiving the highest similarity ratings across categories 5, 4, and 3. Further details can be found in \autoref{appendix: chatgpt_div}.\\

\textbf{For Visualization:} We randomly selected 60 samples from the SST-2 dataset and converted them into vectors using the Sentence-BERT 'all-MPNet-Base-v2' model. We used T-SNE \cite{van2008visualizing} to create a 2D representation of these vectors ((see \autoref{fig:tsne})). TPRL-generated samples closely resemble the original data, demonstrating that small changes in the semantic space can mislead the classifier, revealing vulnerabilities in LM-based classifiers. We consistently observed similar results across different datasets, classifiers, and random samples(refer to \autoref{appendix c} for more details).

\subsubsection{Does TPRL Learned Attacking Policy Universal?}
To investigate this question, we employed a fine-tuned generator to create samples targeting specific classifiers, like BERT-Base. Then, we fine-tuned another classifier, BERT-Large, using these generated samples to assess performance changes, repeating this process for five classifiers. \autoref{unv_policy_table} displays the outcomes in the SST-2 dataset(see \autoref{appendix d} for the remaining datasets), revealing that most classifiers benefited from the samples generated by other classifiers, surpassing the naive baseline. This underscores the universality of the learned attacking policy. Notably, in certain instances, the improvement for the attacked classifier equaled that of the transferred classifiers despite each classifier having distinct errors prior to adversarial training(\autoref{appendix:ceanaly}).\section{Ablation \& Analysis Studies}\label{sec:abstudies}In this section, we conduct ablations and analyze TPRL's key components' impact on baseline improvements.\\
\textbf{TRPL Diversity Enhances Classifier Performance:} We explore diversity's impact on performance, comparing TPRL with the baseline methods. Similarity scores increased with reduced diversity post-adversarial training, while the MI metric sustains diversity. Human evaluations support MI's alignment. Also, we use GPT-3.5 to confirm TPRL's diversity; details in \autoref{appendix: chatgpt_div}.

\textbf{The Importance of Targeted Component}: Our TPRL implementation has a targeted component using RL. However, we evaluated using only the fine-tuned paraphraser without RL (UNTP), which showed inconsistent performance compared to TPRL (\autoref{table:comp_lfs}). Similarly, other baselines requiring a targeted component also yielded minor improvements. TPRL outperformed other models across most metrics (\autoref{table:rf}), emphasizing the importance of the targeted component.

\begin{table*}[htbp]

\begin{center}
\small
\caption{\label{table:gen_samples}Comparing original \& adversarial examples generated by TPRL.}
\label{tbl:co}
\resizebox{\textwidth}{!}{%
\begin{tabular}{@{}l l c c c@{}}
\toprule

\large \textbf{Original:}  i complain all the time about seeing the same ideas repeated in films over and over again, \\ \large	but the bourne identity proves that a fresh take is always possible. \\\\
\large	\textbf{Adversarial:} ive always been complaining that movies repeat the same ideas  but the identity of bourne \\ \large proves that you dont need to reinvent the wheel\\
\midrule

\large	\textbf{Original:} something must be done to halt this terrible wave of immigrants entering britian \\\\
\large	\textbf{Adversarial:} in order to stop a terrible wave of immigration into britian it is necessary to do something\\
\midrule

\large	\textbf{Original:} i live in oslo and race mixing is a huge problem \\\\
\large	\textbf{Adversarial:} theres a huge problem of race mixing in oslo\\
\midrule

\large	\textbf{Original:} william shatner as a pompous professor is the sole bright spot \\\\
\large	\textbf{Adversarial:} the only bright spot of the movie is william shatner in a pompous professor\\
\midrule

\large \textbf{Original:} if you enjoy more thoughtful comedies with interesting conflicted characters; this one is for you \\\\
\large \textbf{Adversarial:} this is a good comedy for those who like thoughtful comedies with interesting characters\\
\midrule
\large \textbf{Original:} any ways smearing a single dna steam with black dna is horrible \\\\
\large \textbf{Adversarial:} any way it seems disgusting to smear one steam with black dna\\
\midrule
\large \textbf{Original:} we spend so much time on black people when the real problem is asians and hispanics \\\\
\large \textbf{Adversarial:} we spend too much time and effort worrying about black while the asians and hispanics were the ones\\

 \bottomrule
\end{tabular}%
}

\normalsize
\end{center}
\end{table*}

\textbf{Number of Generated Samples Impact:} We explore the link between increasing sample size and the resultant performance improvements for both TPRL and a vanilla classifier. TPRL's average sample generation per dataset: SST-2: 359, SST-5: 2993, HS: 135, OFF: 407, AG's News: 3825. We examined the correlation between sample count (normalized by total set size) and TPRL's performance boost, finding no significant correlation (Pearson correlation of 0.144, \( p \ll 0.05 \)). The insignificance stems from performance enhancements' variability due to classifier weaknesses. Thus, we introduced a Mutual Implication filtering criterion of 0.5 for selecting high-quality samples, improving model performance without irrelevant sampling.\\

\textbf{Sample adversarial texts.} \autoref{tbl:co} shows examples of generated samples by TPRL. TPRL introduces changes by paraphrasing the original sentence while preserving the original sentence’s meaning and equivalence. Paraphrasing can be noticed in changing the writing style order of words and substituting or introducing new words simultaneously in one sentence, which is an advantage over word-level attacks that introduce limited operations for one sentence.








\section{Conclusion}
In this paper, we propose TPRL, an adversarial generation method designed to improve classification model robustness. Our innovation involves reinforcement training to learn various attack strategies automatically. We confirm TPRL's effectiveness in four classification tasks, consistently producing high-quality adversarial samples that represent edge cases with minimal distortion in the data distribution.

\section*{Limitations}
One limitation of our work is the reliance on a single scalar reward for optimization, despite the problem having dual objectives: confusion and maintaining similarity. We recommend investigating alternative techniques, such as Multi-objective Reinforcement Learning, to address this limitation. This approach has the potential to enhance performance by optimizing both objectives concurrently. Moreover, the datasets used in paraphrasing currently need longer sequences, approximately 256 tokens, which restricts our approach to generating adversarial samples for longer sequences.

\section*{Ethics Statement}
Enhancing classifier performance is of utmost importance, especially considering the prevalence of hate and offensive speech on social media platforms. Many users attempt to circumvent the classifier's detection capabilities by altering their writing style or incorporating unfamiliar words, thereby creating edge cases where the classifier needs to identify such content accurately. This paper presents an innovative approach to generating these edge cases and leveraging adversarial training to enhance the classifier's ability to detect and protect against such samples.

\section*{Acknowledgements}
The authors would like to thank the members of the Wicked Applied Security Practices (WASP) Lab from the universities of Windsor/Victoria for their support in the annotation process. 

This research is supported by the Vector Scholarship in Artificial Intelligence, provided through the Vector Institute and Natural Sciences and Engineering Research Council of Canada (NSERC) by NSERC Discovery Grant. This research was enabled in part by support provided by Compute Ontario and the Digital Research Alliance of Canada.

\bibliography{anthology,custom}

\clearpage

\onecolumn
\appendix
\section{Natural Language Policy Optimization vs PPO}\label{appendix A}

NLPO (Natural Language Policy Optimization) is proposed to address the problem of large action spaces in language generation tasks. \cite{ramamurthy2022reinforcement} showed that existing RL algorithms struggle with these spaces, as seen in models like GPT-2/3 and T5 with their extensive vocabularies of 50K and 32K, respectively, and even more prominent with the recent models. NLPO introduces a masking policy that is periodically updated and applies a top-p sampling technique to mask out irrelevant tokens during training. This helps balance the inclusion of task-relevant information with the risk of reward hacking. By extending the PPO algorithm, NLPO aims to improve the stability and effectiveness of training language models. NLPO achieves that by NLPO utilizing top-p sampling via generating, which limits the token selection to the smallest set where the cumulative probability exceeds a given threshold parameter p \cite{holtzman2018learning}.

\section{Implementation Details For Baseline Methods}\label{appendix B}

We utilized the codebase provided by the authors for the baseline methods. Nevertheless, certain aspects were not explicitly addressed in their paper or the baseline implementation. Despite this, we made efforts to adapt these aspects in order to ensure minimal disruption to the overall framework.
\subsection{SCPN}

SCPN is an approach that leverages an LSTM model trained on a large back-translation corpus to generate paraphrases. These paraphrases are then parsed using the Stanford parser. In order to generate adversarial samples, SCPN employs ten different parsing templates. We adopted the same methodology as the authors by utilizing their pre-trained models and following their steps to generate parse trees for our datasets using the Stanford parser \cite{manning2014stanford}. However, the paper and codebase did not provide details on how to select the appropriate parsing template.
We devised a strategy for choosing the most suitable parsing template to address this. Given that the model generates ten templates, we initially use the pre-trained model to generate multiple paraphrases of input "x" using each template. Subsequently, we individually query the victim model with each generated paraphrase. We then measure the confusion and mutual implication score for each paraphrase and select the sample that yields the highest scores as the chosen paraphrase. This process ensures we prioritize the paraphrase that maximizes confusion and mutual implication with the victim model.
\subsection{StyleAdv}

StyleAdv is an approach that leverages the power of STRAP (Style Transfer via Paraphrasing), a style transfer framework. This approach incorporates five distinct style transfer models, namely Bible, Poetry, Shakespeare, Lyrics, and Tweets, each capable of generating a unique style.
To ensure consistency and reproducibility, we meticulously followed the procedure outlined in the paper and codebase for the adversarial generation of our datasets. However, we encountered a missing reference to the similarity model in the paper and codebase. Upon contacting the authors, they informed us that any similarity model would suffice. Consequently, we opted for the "all-MPNet-Base-v2" model, renowned for its exceptional performance on the semantic textual similarity (STS) benchmark \cite{muennighoff2022mteb}. We employed this model to measure cosine similarity, a reliable metric for comparing sentence similarity.
The adversarial generation process unfolds: utilizing each style transfer model, we generate ten paraphrases for a single sentence, resulting in 50 paraphrases. Subsequently, we subject these generated paraphrases to classification by our classifier, measuring both the confusion and cosine similarity. If multiple examples cause the classifier to produce incorrect outputs, we select the adversarial example with the highest cosine similarity to the original input as the final choice.
\section{Visualization Results}\label{appendix c}
We utilized the T-SNE technique to generate a two-dimensional representation of the vectorized samples. To obtain these vectors, we employed the "all-MPNet-Base-v2" model and selected 60 random samples from each class for each dataset. We tried various numbers of samples, and 60 gives a clear observation of the phenomena. 

 Subsequently, T-SNE was applied to obtain the two-dimensional representation. Upon examining the figures depicting the proposed datasets, the same observation for TPRL-generated samples in SST-2 holds for the other datasets, which is the samples generated by TPRL displayed a striking resemblance to the original data, with instances of overlapping or partial overlap with the original sentences. This intriguing behavior sheds light on a crucial finding: even a minor shift in the semantic space relative to the original sentence can yield a sentence that successfully deceives the classifier.
\subsection{SST-2}
\begin{figure}[ht] 
  \begin{subfigure}[b]{0.5\linewidth}
    \centering
    \fbox{\includegraphics[width=0.9\linewidth]{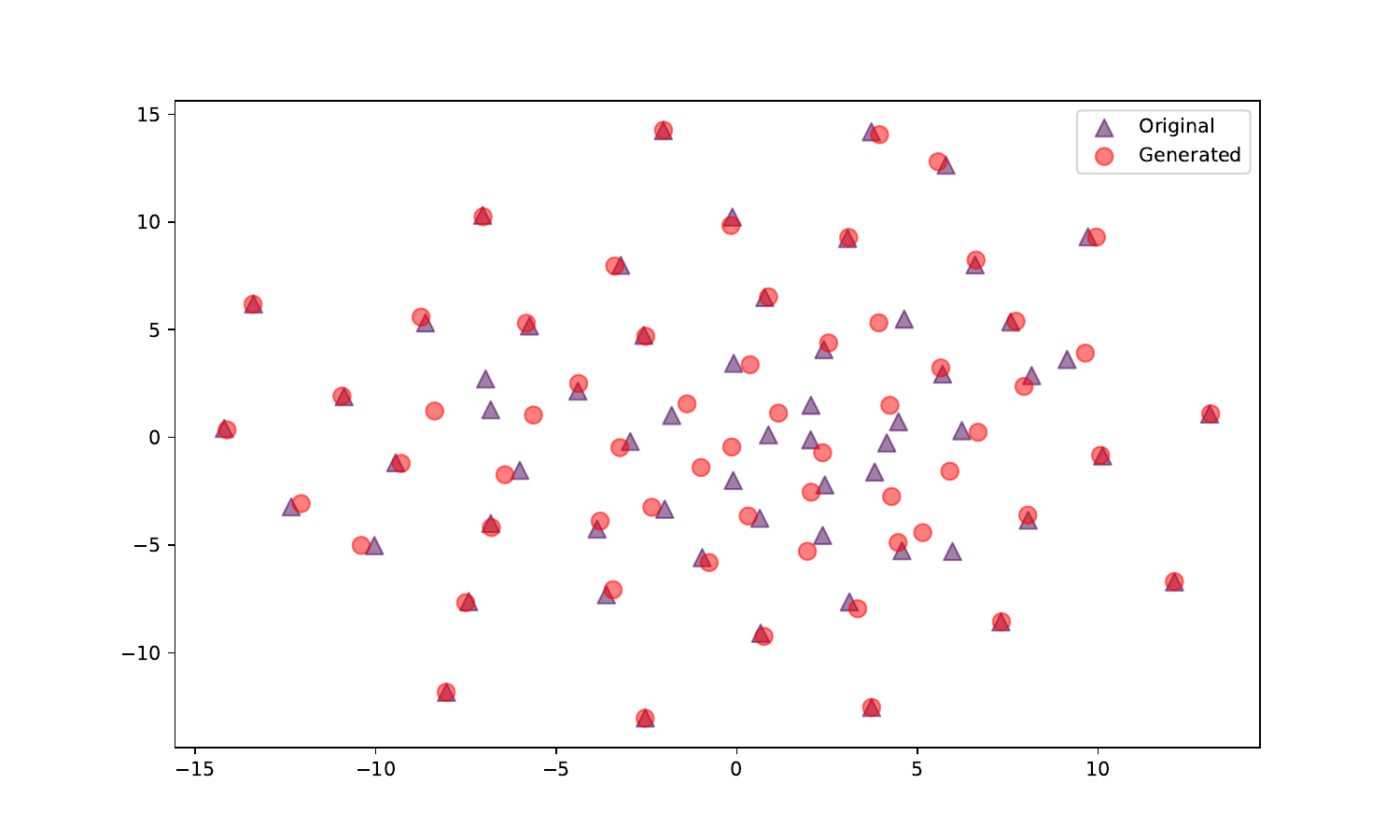}}
    \caption{Class-0} 
    \label{fig7:a} 
    \vspace{4ex}
  \end{subfigure}
  \begin{subfigure}[b]{0.5\linewidth}
    \centering
    \fbox{\includegraphics[width=0.9\linewidth]{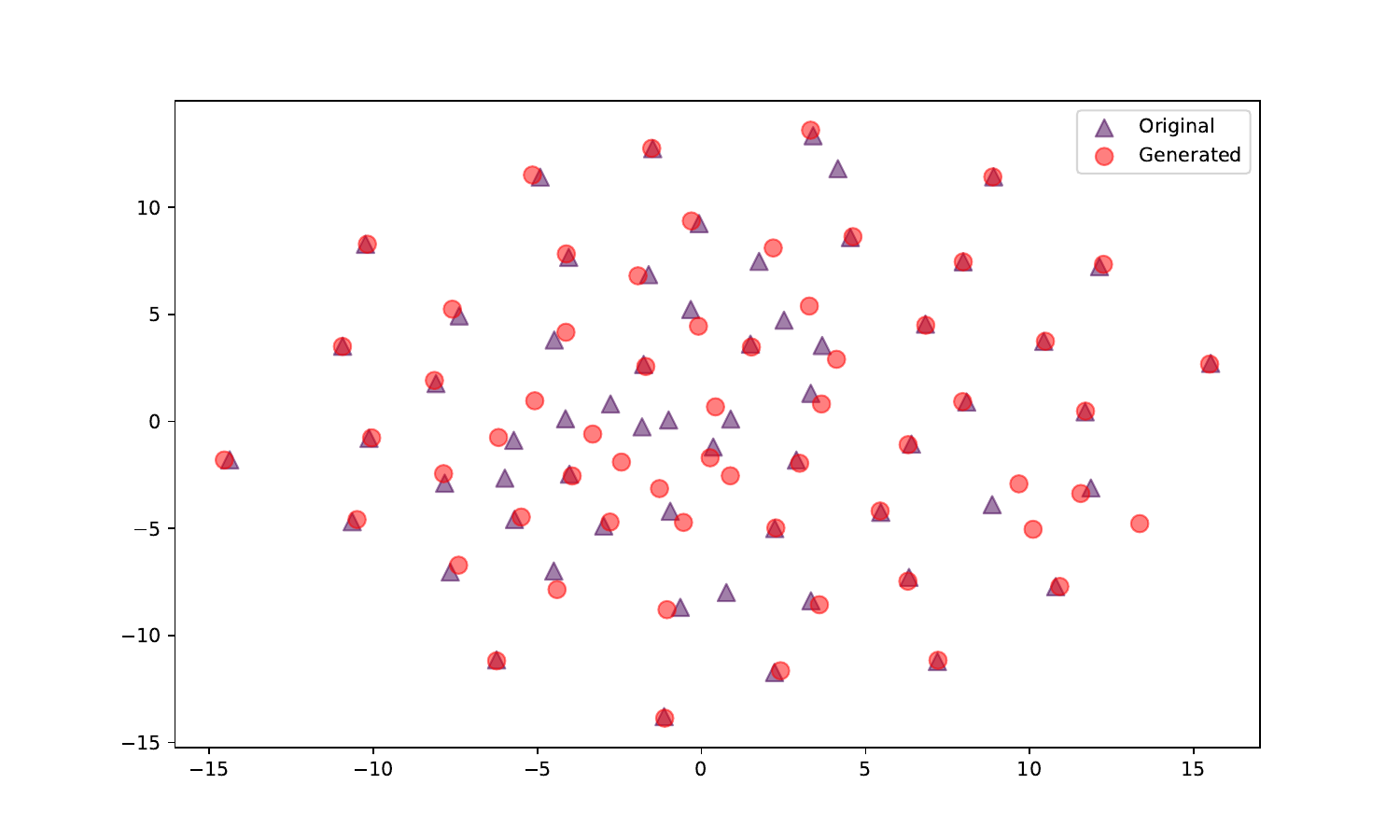}}
    \caption{Class-1} 
    \label{fig7:b} 
    \vspace{4ex}
  \end{subfigure} 
  \caption{T-SNE visualization of the vectorized original and TPRL-adversarial sentences in the SST-2. The adversarial sentences (circles) mostly overlap with the original sentences (triangles), suggesting that generated sentences maintain the original class distribution.}
  \label{fig:sp_125m} 
\end{figure}

\subsection{SST-5}
\begin{figure}[ht] 
  \begin{subfigure}[b]{0.38\linewidth}
    \centering
    \fbox{\includegraphics[width=0.9\linewidth]{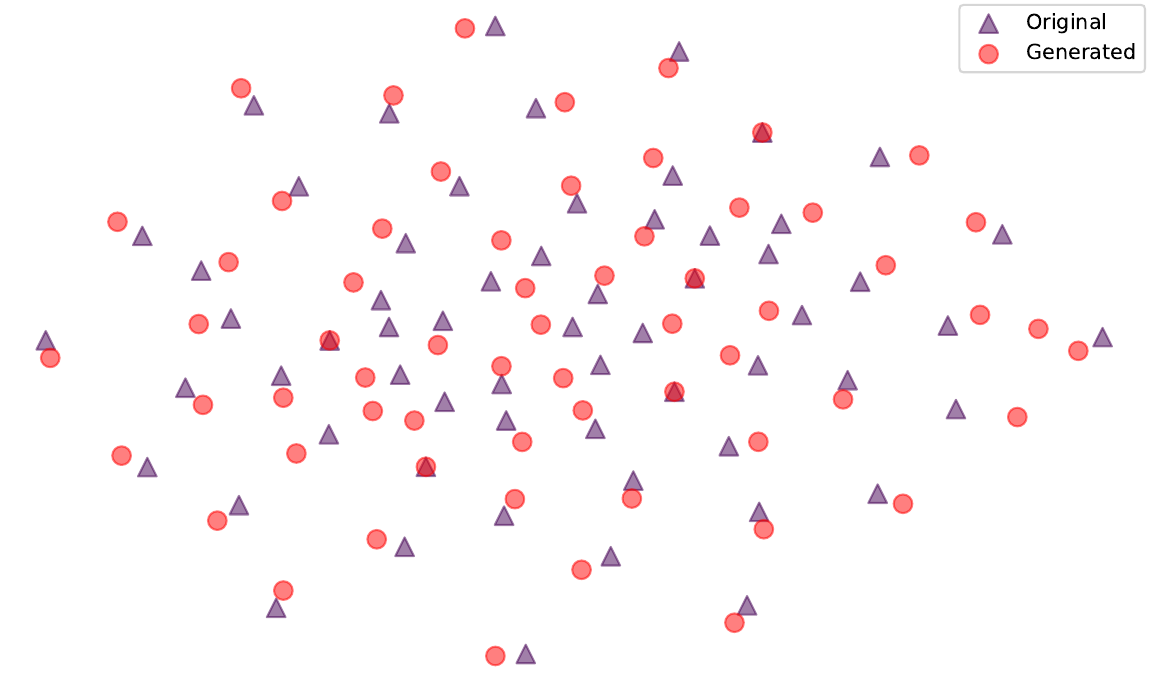}}
    \caption{Class-0-Train set} 
    \label{fig7:a} 
    \vspace{4ex}
  \end{subfigure}
  \begin{subfigure}[b]{0.38\linewidth}
    \centering
    \fbox{\includegraphics[width=0.9\linewidth]{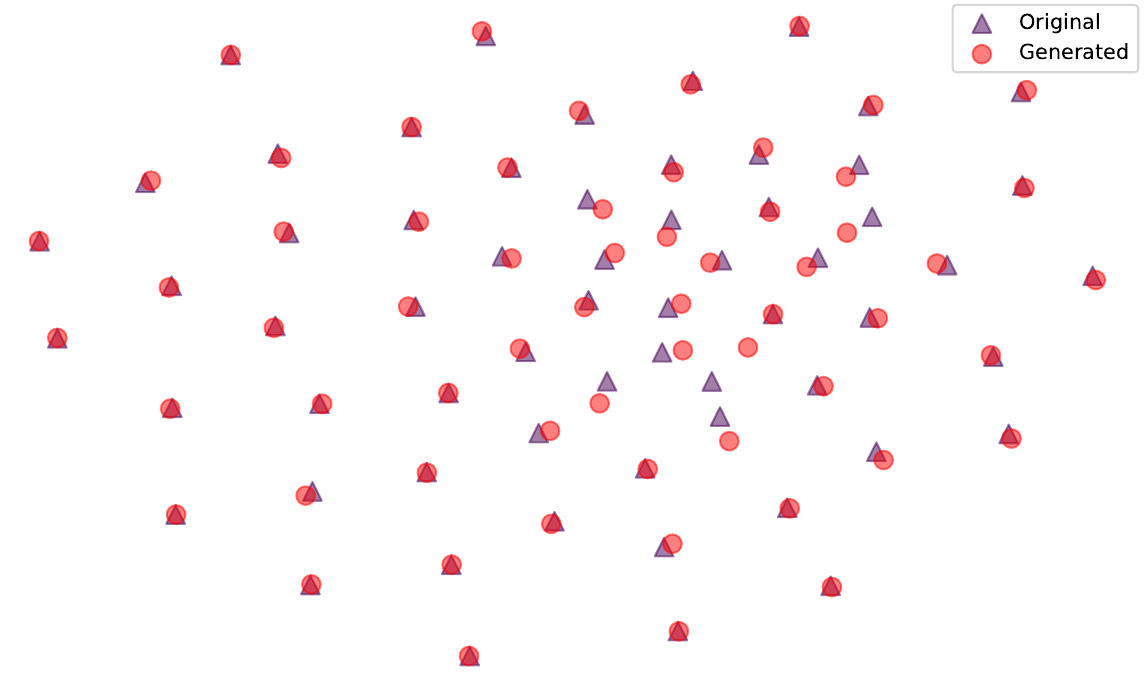}}
    \caption{Class-1-Train set} 
    \label{fig7:b} 
    \vspace{4ex}
  \end{subfigure} 
  \begin{subfigure}[b]{0.38\linewidth}
    \centering
    \fbox{\includegraphics[width=0.9\linewidth]{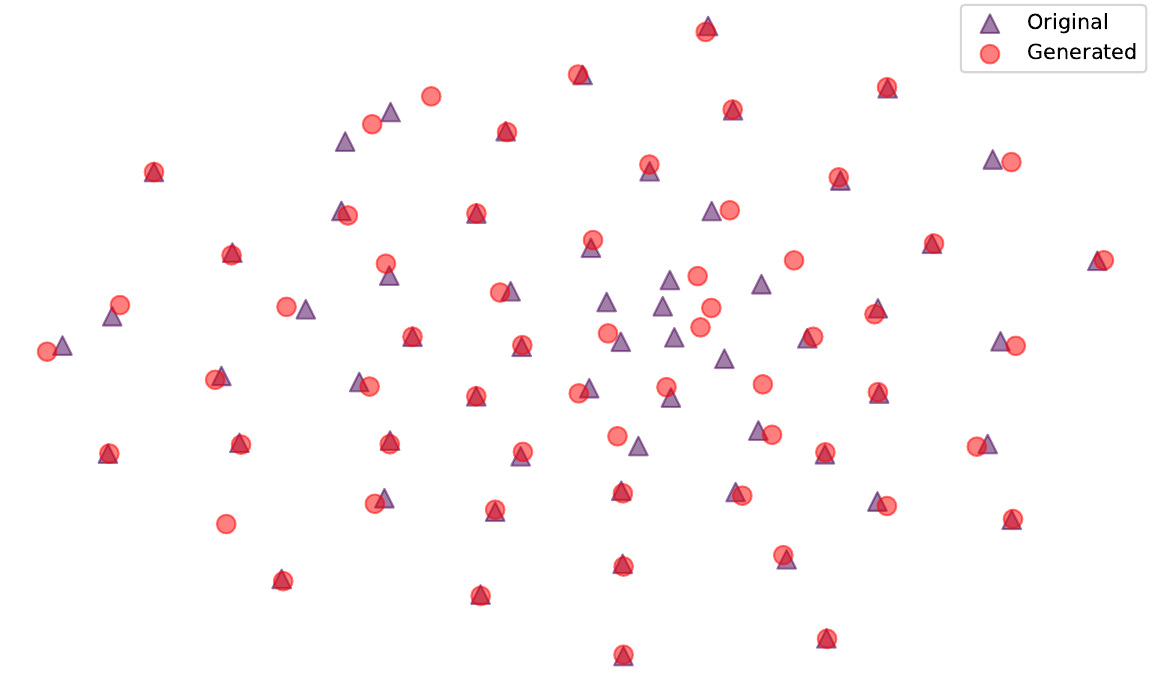}} 
    \caption{Class-0-Test set} 
    \label{fig7:c} 
  \end{subfigure}
  \begin{subfigure}[b]{0.38\linewidth}
    \centering
    \fbox{\includegraphics[width=0.9\linewidth]{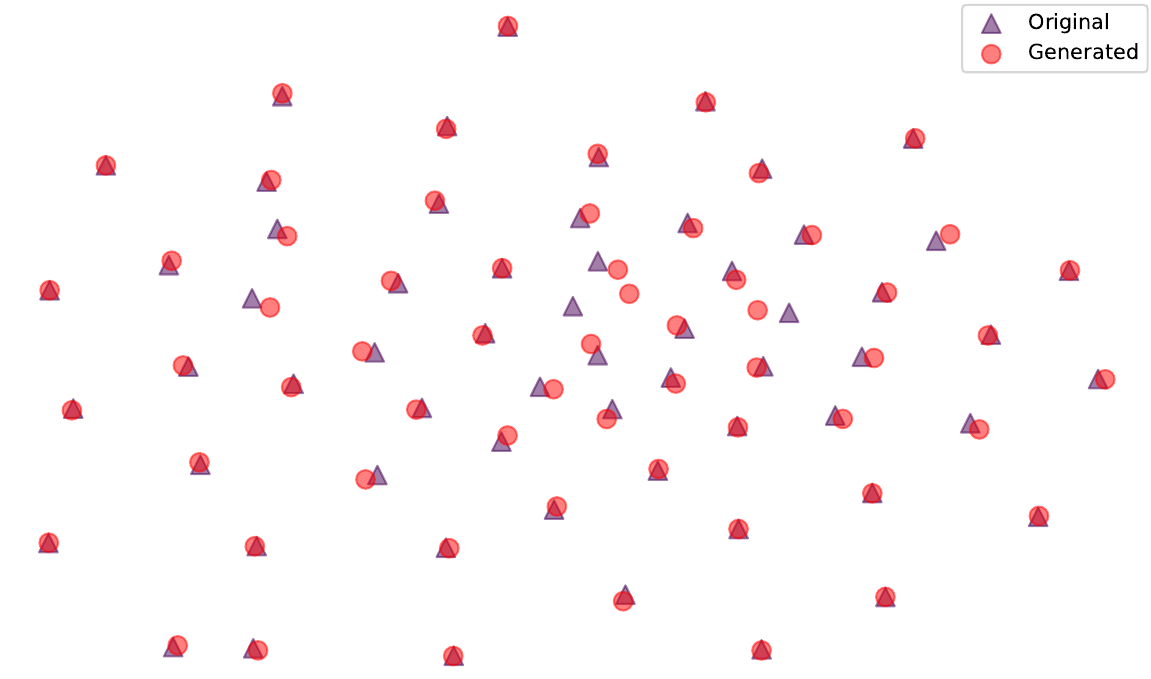}}
    \caption{Class-1-Test set} 
    \label{fig:sads} 
  \end{subfigure} 
  \caption{T-SNE visualization of the vectorized original and TPRL-adversarial sentences in the SST-5. The adversarial sentences (circles) mostly overlap with the original sentences (triangles), suggesting that generated sentences maintain the original class distribution.}
  \label{fig:sp_125m} 
\end{figure}
\clearpage
\subsection{Offensive Dataset}
\begin{figure}[ht] 
  \begin{subfigure}[b]{0.5\linewidth}
    \centering
    \fbox{\includegraphics[width=0.9\linewidth]{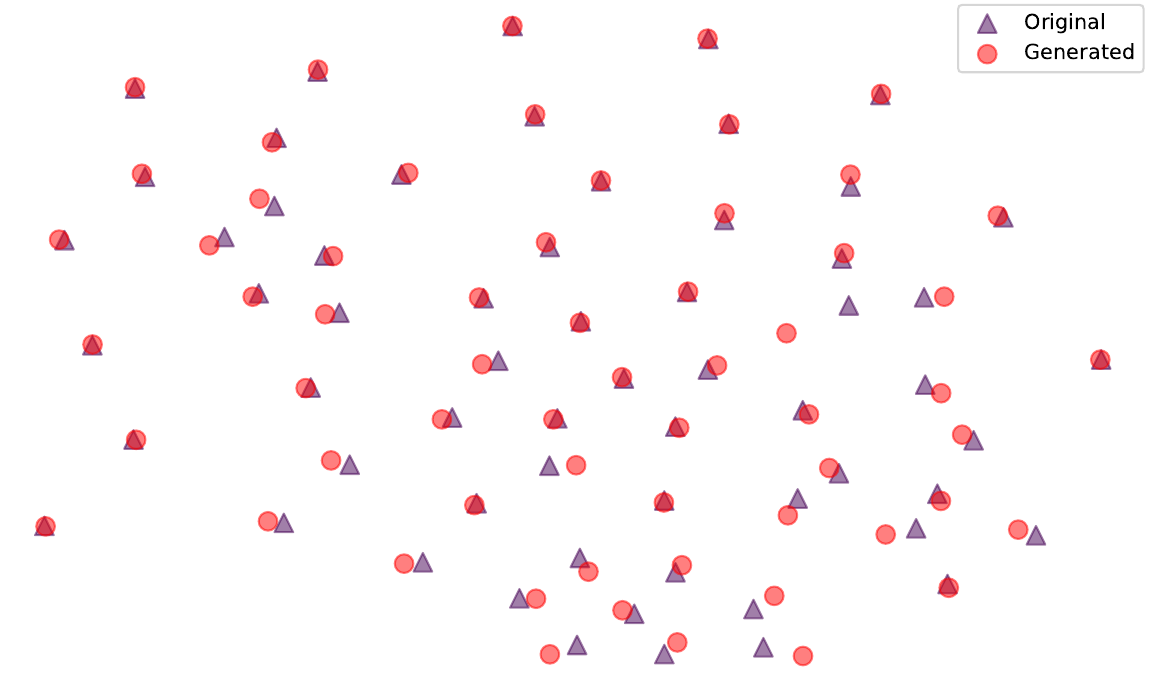}}
    \caption{Class-0} 
    \label{fig7:a} 
    \vspace{4ex}
  \end{subfigure}
  \begin{subfigure}[b]{0.5\linewidth}
    \centering
    \fbox{\includegraphics[width=0.9\linewidth]{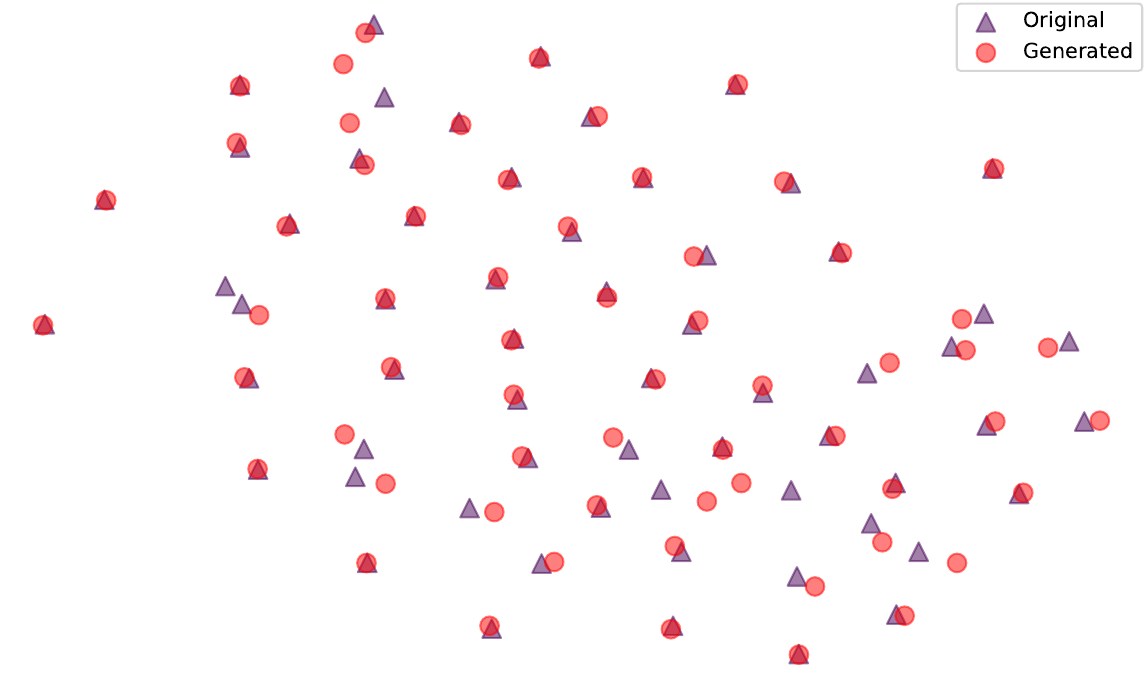}}
    \caption{Class-1} 
    \label{fig7:b} 
    \vspace{4ex}
  \end{subfigure} 
  \caption{T-SNE visualization of the vectorized original and TPRL-adversarial sentences in OFF. The adversarial sentences (circles) mostly overlap with the original sentences (triangles), suggesting that generated sentences maintain the original class distribution.}
  \label{fig:sp_125m} 
\end{figure}

\subsection{Hate Dataset}
\begin{figure}[ht] 
  \begin{subfigure}[b]{0.5\linewidth}
    \centering
    \fbox{\includegraphics[width=0.9\linewidth]{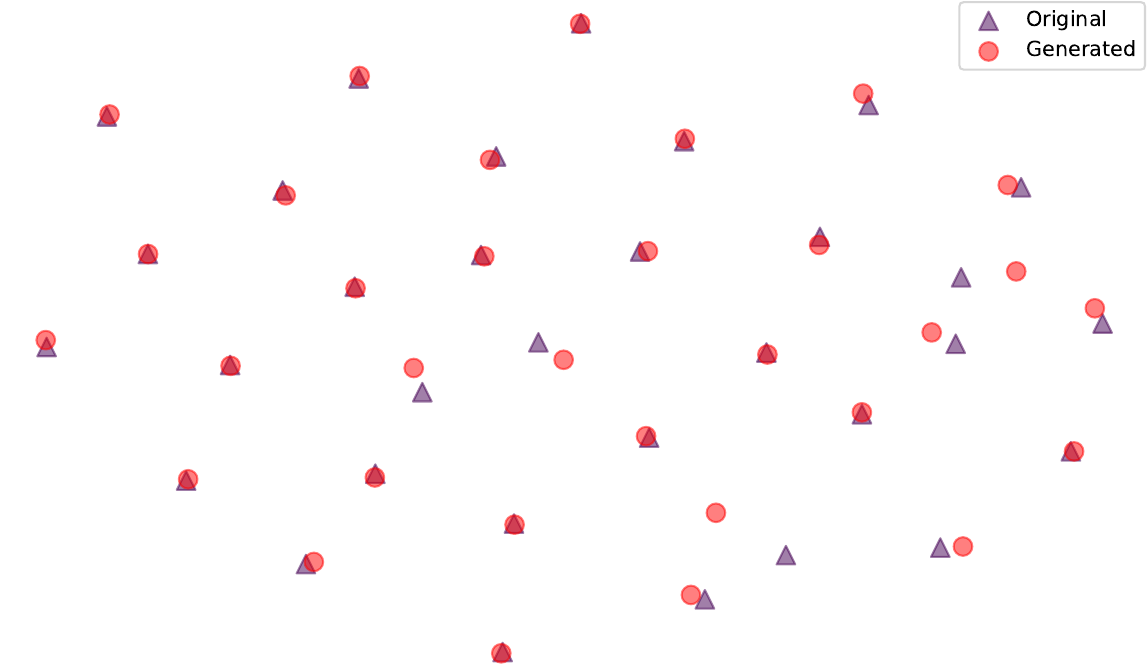}}
    \caption{Class-0} 
    \label{fig7:a} 
    \vspace{4ex}
  \end{subfigure}

  \caption{T-SNE visualization of the vectorized original and TPRL-adversarial sentences in HATE. The adversarial sentences (circles) mostly overlap with the original sentences (triangles), suggesting that generated sentences maintain the original class distribution.}
  \label{fig:sp_125m} 
\end{figure}
\clearpage
\subsection{Agnews Dataset}

\begin{figure}[ht] 
  \begin{subfigure}[b]{0.5\linewidth}
    \centering
    \fbox{\includegraphics[width=0.9\linewidth]{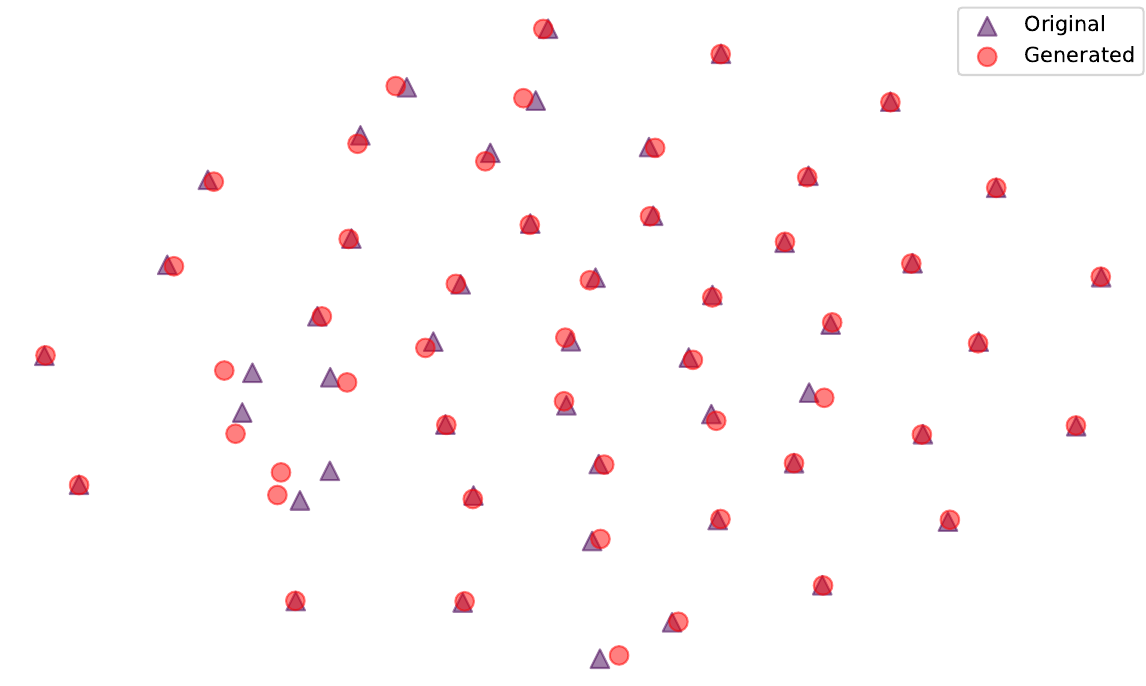}}
    \caption{Class-0} 
    \label{fig7:a} 
    \vspace{4ex}
  \end{subfigure}
  \begin{subfigure}[b]{0.5\linewidth}
    \centering
    \fbox{\includegraphics[width=0.9\linewidth]{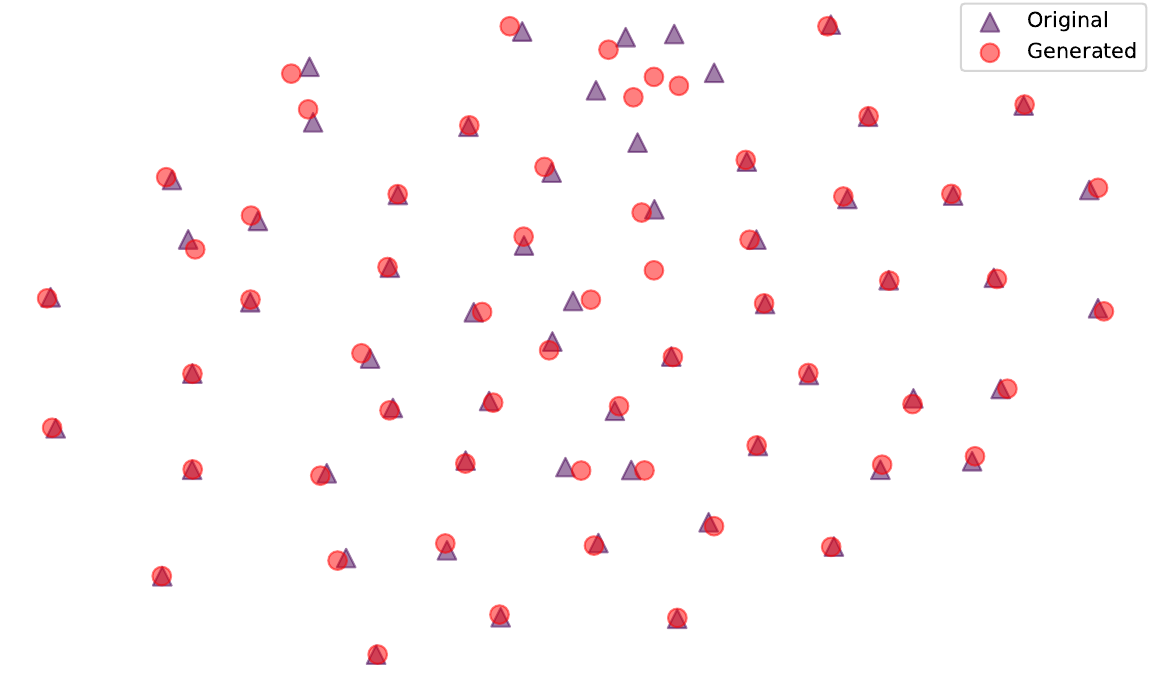}}
    \caption{Class-1} 
    \label{fig7:b} 
    \vspace{4ex}
  \end{subfigure} 
  \begin{subfigure}[b]{0.5\linewidth}
    \centering
    \fbox{\includegraphics[width=0.9\linewidth]{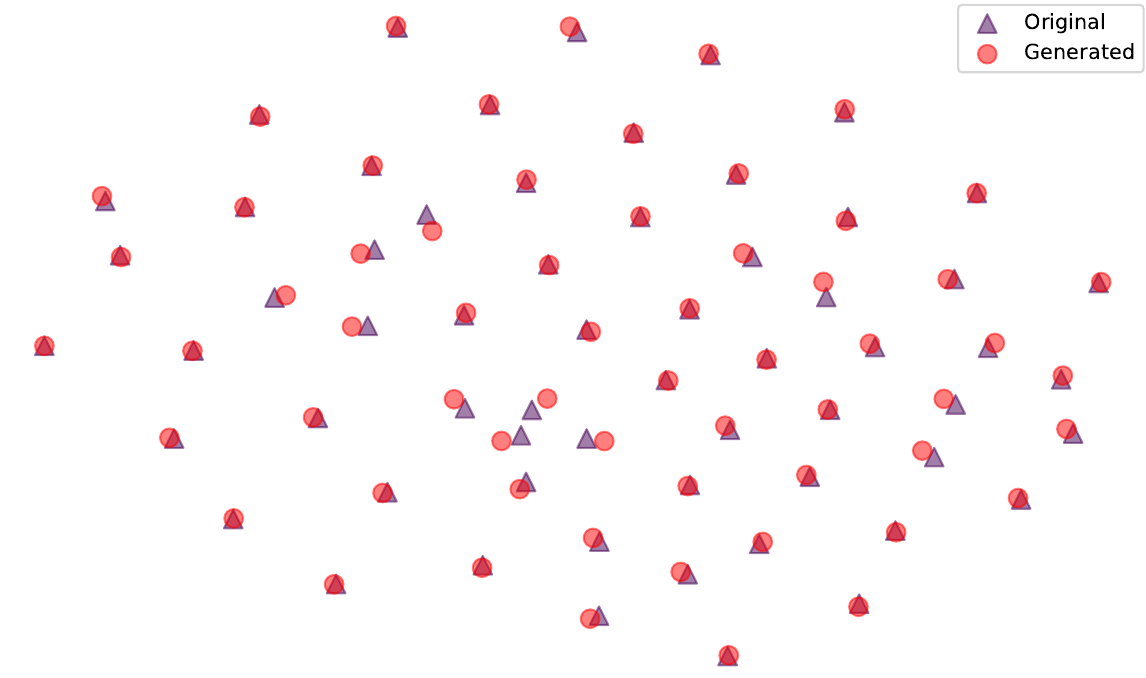}} 
    \caption{Class-2} 
    \label{fig7:c} 
  \end{subfigure}
  \begin{subfigure}[b]{0.5\linewidth}
    \centering
    \fbox{\includegraphics[width=0.9\linewidth]{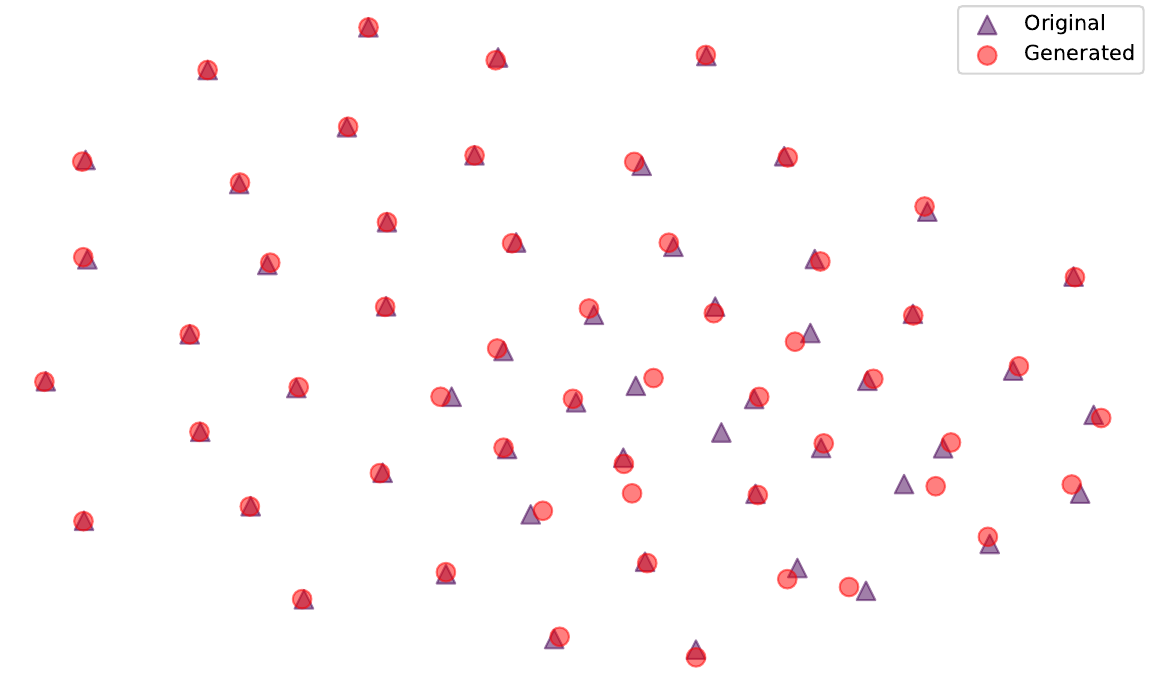}}
    \caption{Class-3} 
    \label{fig:sads} 
  \end{subfigure} 
  \caption{T-SNE visualization of the vectorized original and TPRL-adversarial sentences in AG's News. The adversarial sentences (circles) mostly overlap with the original sentences (triangles), suggesting that generated sentences maintain the original class distribution.}
  \label{fig:sp_125m} 
\end{figure}
\clearpage
\section{Universal Policy}\label{appendix d}
As previously discussed, TPRL's learned policy demonstrates remarkable universality across multiple datasets and classifiers. In this section, we extensively analyze the learned policy's performance on each dataset, specifically focusing on its efficacy with four different classifiers. The following results highlight the consistent and impressive performance of the learned policy across diverse datasets and classifiers.

\subsection{SST-5}

\begin{table*}[htbp]
\begin{center}
    \centering
\resizebox{16cm}{!}{%
\begin{tabular}{c|ccccc}
\hline

\multirow{2}{*}{\textbf{Policy/Classifier}}& 
\textbf{BERT$_{Base}$} & \textbf{BERT$_{Large}$} & \textbf{RoBERTa$_{Base}$} & \textbf{RoBERTa$_{Large}$} 
& \textbf{DeBERTa-v3$_{Large}$} \\
  & (\%) & (\%) & (\%) & (\%) & (\%)\\
\hline
None & 53.52 & 53.52 & 54.79 & 58.30 & 58.46
\\
Policy-BERT$_{Base}$ & 53.52 & 54.88 & \bf 56.42 & $\star$& 57.10
\\
Policy-BERT$_{Large}$ & 52.89 & 54.93 & 55.61& $\star$& 
59.00
\\
Policy-RoBERTa$_{Base}$ & 51.62 &  54.84 & 56.15& $\star$ & \bf 59.54
\\
Policy-RoBERTa$_{Large}$ &  \bf 54.07 & \bf 55.15 & $\star$ & \bf 58.95 & 58.91
\\
Policy-DeBERTa-v3$_{Large}$ & 53.52 & 54.61 & 55.56 & $\star$ & 58.77
\\
\hline
\end{tabular}
}
\caption{\label{unv_policy_table_sst5}
Accuracy Results of Different Classifiers Trained With The Examples Generated By Various Attacking Policies On The SST-5 Dataset. Showing the Universal Policy. The best comparable performances policy for the classifier is bolded}

\end{center}

\end{table*}

\subsection{Offensive Dataset}
\begin{table*}[htbp]
\begin{center}
    \centering
\resizebox{15cm}{!}{%
\begin{tabular}{c|ccccc}
\hline

\multirow{2}{*}{\textbf{Policy/Classifier}}& 
\textbf{BERT$_{Base}$} & \textbf{BERT$_{Large}$} & \textbf{RoBERTa$_{Base}$} & \textbf{RoBERTa$_{Large}$} 
& \textbf{DeBERTa-v3$_{Large}$} \\
  & (\%) & (\%) & (\%) & (\%) & (\%)\\
\hline
None & 84.76 & 85.11 & 83.95 & 85.93 & 84.65
\\
Policy-BERT$_{Base}$ & 85.11 & 85.00 & 83.95 & 72.09& 84.76
\\
Policy-BERT$_{Large}$ & 85.11 & 85.58 &84.76& 84.76& 
84.53
\\
Policy-RoBERTa$_{Base}$ & 85.11 &  84.88 & 85.00& \bf 85.93 & 84.53
\\
Policy-RoBERTa$_{Large}$ & 84.30 & \bf 85.93 & \bf 85.23 & 84.53 & \bf 86.04
\\
Policy-DeBERTa-v3$_{Large}$ & \bf 85.81 & 85.81 & \bf 85.23 & 83.02 & 85.93
\\
\hline
\end{tabular}
}
\caption{\label{unv_policy_table_off}
Accuracy Results of Different Classifiers Trained With The Examples Generated By Various Attacking Policies On The OFF Dataset. Showing the Universal Policy. The best comparable performances policy for the classifier is bolded}

\end{center}

\end{table*}

\subsection{Hate Dataset}
\begin{table*}[htbp]
\begin{center}
    \centering
\resizebox{13cm}{!}{%
\begin{tabular}{c|ccccc}
\hline

\multirow{2}{*}{\textbf{Policy/Classifier}}& 
\textbf{RoBERTa$_{Base}$} & \textbf{RoBERTa$_{Large}$} 
& \textbf{DeBERTa-v3$_{Large}$} \\
  & (\%) & (\%) & (\%)\\
\hline
None & 91.90 & \bf 92.45 & 89.50 
\\
Policy-RoBERTa$_{Base}$ & 92.45 &  91.75 & \bf 93.80
\\
Policy-RoBERTa$_{Large}$ & \bf 92.75 & 92.05 & 92.60 
\\
Policy-DeBERTa-v3$_{Large}$ & 90.65 & 91.45 & 92.30
\\
\hline
\end{tabular}
}
\caption{\label{unv_policy_table_hs}
Accuracy Results of Different Classifiers Trained With The Examples Generated By Various Attacking Policies On The HATE Dataset. Showing the Universal Policy. The best comparable performances policy for the classifier is bolded}

\end{center}

\end{table*}

\clearpage
\subsection{AG's News Dataset}
\begin{table*}[htbp]
\begin{center}
    \centering
\resizebox{16cm}{!}{%
\begin{tabular}{c|ccccc}
\hline

\multirow{2}{*}{\textbf{Policy/Classifier}}& 
\textbf{BERT$_{Base}$} & \textbf{BERT$_{Large}$} & \textbf{RoBERTa$_{Base}$} & \textbf{RoBERTa$_{Large}$} 
& \textbf{DeBERTa-v3$_{Large}$} \\
  & (\%) & (\%) & (\%) & (\%) & (\%)\\
\hline
None & 93.97 & 94.18 & 93.78 & 93.92 & 93.94
\\
Policy-BERT$_{Base}$ & \bf 94.46 & \bf 94.60 & 93.92 & 93.97& \bf 94.23
\\
Policy-BERT$_{Large}$ & 94.21 & 94.36 & 93.73 & 94.02 & 
93.85
\\
Policy-RoBERTa$_{Base}$ & 94.17 & 94.28 & 93.93 & 93.78 & 91.01
\\
Policy-RoBERTa$_{Large}$ & 93.85 & 94.17 & 93.72 & \bf 94.21 & 94.13
\\
Policy-DeBERTa-v3$_{Large}$ & 94.13 & 94.26 & \bf 93.94 & 93.67 & 94.22
\\
\hline
\end{tabular}
}
\caption{\label{unv_policy_table_ag}
Accuracy Results of Different Classifiers Trained With The Examples Generated By Various Attacking Policies On The AG's News Dataset. Showing the Universal Policy. The best comparable performances policy for the classifier is bolded}

\end{center}

\end{table*}

\section{Automatic Evaluation}\label{appendix e}
We adopted a comprehensive multi-perspective methodology to assess the quality of the generated adversarial samples, ensuring the following factors were taken into consideration: fluency, as determined by Perplexity (PPL) scores obtained from the GPT-2-XL language model \cite{radford2019language}. However, recognizing the inherent limitations of perplexity in accurately evaluating fluency, we supplemented this metric with the accuracy of a RoBERTa-large classifier, which was trained on the CoLA corpus \cite{warstadt2019neural}. This classifier offers valuable insights into the grammatical acceptability of the generated samples. For measuring similarity, we utilized the "all-MPNet-Base-v2" embedding-based SIM model \cite{song2020mpnet, reimers2019sentence} to measure the semantic similarity between the input sentence and the generated samples. This model has demonstrated exceptional performance on the semantic textual similarity (STS) benchmark \cite{muennighoff2022mteb}, making it an ideal choice for our task. To further enhance our evaluation, we also integrated the mutual implication (MI) metric, which effectively captures the inferential role semantics. By incorporating the MI metric, we overcome the limitations of STS in fully capturing the inferential semantics, thereby providing a more comprehensive evaluation of the generated samples. The results for each dataset with each classifier are shown in the following table.
\begin{table*}[h]
\centering
\resizebox{16cm}{!}{%
\begin{tabular}{c|c|cccc|cccc|cccc|cccc|cccc}
\hline
\multirow{2}{*}{\textbf{Dataset}} & Classifier & \multicolumn{4}{c}{$\text{BERT}_{\text{Base}}$} & \multicolumn{4}{c}{$\text{BERT}_{\text{Large}}$} & \multicolumn{4}{c}{$\text{RoBERTa}_{\text{Base}}$} & \multicolumn{4}{c}{$\text{RoBERTa}_{\text{Large}}$} & \multicolumn{4}{c}{$\text{DeBERTa}_{\text{Large}}$}\\
\cline{2-22}
 & Attacker & PPL$\downarrow$&FL$\uparrow$&SIM$\uparrow$&MI$\uparrow$ & PPL$\downarrow$&FL$\uparrow$&SIM$\downarrow$&MI$\uparrow$ & PPL$\downarrow$&FL$\uparrow$&SIM$\uparrow$&MI$\uparrow$ & PPL$\downarrow$&FL$\uparrow$&SIM$\uparrow$&MI$\uparrow$ & PPL$\downarrow$&FL$\uparrow$&SIM$\uparrow$&MI$\uparrow$\\
\hline
\multirow{3}{*}{SST-2} 
 & SCPN 
 & 467.84 & 58.47 & 72.98& 73.20
 & 461.367 & 58.68 & 73.08 & 73.49 
 & 442.519 &  57.29 & 71.78 & 70.67 
 & 442.95 & 58.23 & 71.95 & 70.16 
 & 434.443 & 57.88 & 71.99 & 70.80\\
 
 & StyleAdv 
 & 1114.599 & 58.29 & \bf 76.29 & 57.27
 & 1173.938 & 57.05 & \bf 74.57 & 53.44 
 & 1183.201& 56.75 & \bf 74.24 & 52.96
 & 540.362 & 56.38 & \bf 73.93& 51.35 
 & 1281.874 & 56.49& \bf 73.33 & 50.31\\
 
  & TPRL 
  & \bf 293.436 & \bf 88.97 & 67.18 & \bf 86.11 
  & \bf 327.538 & \bf 86.36 & 72.21&  \bf 87.18 
  & \bf 396.76 & \bf 87.59 & 72.94 & \bf 85.02 
  & \bf 405.21 & \bf 86.58 & 58.14&  \bf 90.11 
  & \bf 438.16 & \bf 87.0 & 70.81 & \bf 85.8 \\
\hline

\multirow{3}{*}{SST-5} 
 & SCPN 
 & 462.351 & 59.22 & 77.70 & 83.66
 & 444.079 & 59.10 & 78.23 & 84.19 
 & 443.576 & 59.10 & 78.12 & 84.04 
 & 430.548 & 58.81 & 78.21 & 84.06 
 & 461.905 & 59.00&  77.09 & 82.13\\
 
 & StyleAdv 
 & \bf 257.311 & 68.24 & \bf 90.43 & 78.98 
 & \bf 256.632 & 67.30 & \bf 89.79 & 77.45 
 & \bf 248.617 & 70.50 & \bf 90.95 & 79.97 
 & \bf 315.372 & 67.70 & \bf 89.49 & 77.26 
 & \bf 261.436 & 69.42 & \bf 89.90 & 78.27\\
 
  & TPRL  
  & 354.96 & \bf 85.94 & 75.2 & \bf 87.53
  & 283.46 & \bf 86.84 & 75.17 &  \bf 85.31  
  & 401.87 & \bf 85.48 & 78.96 & \bf 87.49
  & 319.86 & \bf 86.86 & 74.89 & \bf 85.02
  & 373.00 & \bf 86.72 & 75.90 & \bf 86.22 \\
\hline

\multirow{3}{*}{HS} 
 & SCPN 
 & $\star$ & $\star$ & $\star$ & $\star$
 & $\star$ & $\star$ & $\star$ & $\star$
 & 736.244 & 58.85 & 75.83 & 85.91
 & 739.991 & 58.59 & 75.70 & 85.69 
 & 746.287 & 58.42 & 75.85 & 85.42\\
 
 & StyleAdv 
 & $\star$ & $\star$ & $\star$ & $\star$
 & $\star$ & $\star$ & $\star$ & $\star$ 
 & \bf 367.99  & 57.34 & \bf 79.14 & 60.97 
 & 360.303 & 58.84 & \bf 76.42 & 56.77 
 & 407.045 & 59.15&  \bf 79.22 & 60.02\\
 
  & TPRL 
  & $\star$ & $\star$ & $\star$ & $\star$
  & $\star$ & $\star$ & $\star$ & $\star$
  & 973.77 & \bf 80.61 & 59.14 & \bf 92.15  
  & \bf 212.95 & \bf 89.69 & 74.8 & \bf 89.76 
  & \bf 197.68 & \bf 90.05 & 75.43 & \bf 90.52 \\
\hline

\multirow{3}{*}{OFF} 
 & SCPN 
 & 780.052 & 47.44 & 73.43 & 79.51 
 & 801.278 & 47.46 & 73.39 & 80.21
 & 827.18  & 47.41 & 73.74 & 80.32 
 & 790.174 & 47.58 & 73.61 & 80.40 
 & 817.611 & 47.20 & 73.93 & 80.81\\
 
 & StyleAdv 
 & 1273.684 & 51.28 & 77.12 & 58.73 
 & 2069.405 & 52.09 & 77.45 & 58.48 
 & 1316.525 & 51.91 & 76.64 & 57.20 
 & 1298.085 & 52.84 & 77.46 & 58.27 
 & 1827.177 & 51.87 & 79.25 & 59.90\\
 
  & TPRL 
  & \bf 316.61 & \bf 87.28 & 72.83 & \bf 87.50 
  & \bf 472.94 & \bf 87.17 & 72.47 & \bf 88.58 
  & \bf 412.41 & \bf 85.88 & 70.87 & \bf 87.85
  & \bf 327.49 & \bf 85.69&  67.17 & \bf 90.60
  & \bf 355.65 & \bf 86.73 & 71.13 & \bf 87.32
 \\
\hline

\multirow{3}{*}{AG’s News} 
 & SCPN 
 & 383.066 & 29.05 & 72.62 & 80.76 
 & \bf 382.20 & 29.05 & 72.60 & 80.85 
 & 367.79 & 28.91 & 72.68 & 80.50
 & 371.46 & 28.95 & 72.69 & 80.64 
 & \bf 343.28 & 28.94 & 72.56 & 80.39\\
 
 & StyleAdv 
 & \bf 356.54 & 51.55 & 53.34 & 39.45 
 & 435.91 & 50.70 & 54.92 & 42.15 
 & 425.10 & 49.81 & 57.34 & 43.86 
 & 384.90 & 50.93 & 54.25 & 40.42 
 & 451.27 & 51.55 & 52.88 & 38.78\\
 
  & TPRL 
  & 508.63 & \bf 86.19 & \bf 79.05 & \bf 98.46 
  & 406.12 & \bf 87.02 & \bf 75.68 & \bf 98.23 
  & \bf 277.79 & \bf 88.39 & \bf 74.21 & \bf 97.44 
  & \bf 248.27 & \bf 88.11 & \bf 79.20 & \bf 97.57
  & 600.80 & \bf 85.88 & \bf 76.24 & \bf 98.51\\
\hline
\end{tabular}
}
\end{table*}

\section{Hyperparameters Details}\label{appendix hyper}
The model was trained for thirty epochs as we tried a range of epochs and picked up the best value that achieves a higher reward in the training environment. While a batch size of 32 was chosen empirically, as changing batch size did not affect performance. The training process employed the Lion optimizer, as proposed by \cite{chen2023symbolic}, in their work on symbolic optimization. With a learning rate of $4.9$ $\times$ $10^{-6}$ suggested by \cite{vonwerra2022trl}, the Lion optimizer demonstrated superior convergence compared to the commonly used Adam optimizer \cite{kingma2014adam}.

\section{GPT-3.5 Annotation Details}\label{appendix: chatgpt_div}

\subsection{Measuring Similarity Via GPT-3.5}
To assess the similarity of larger generated samples, we used GPT-3.5. Using “from 1 to 5, how much is the generated sentence similar to the original (1 being very dissimilar and 5 being very similar)?” as a prompt. We randomly selected 100 samples from each framework in the SST-2 dataset. Ratings were given on a scale of 1 to 5, with 1 being very dissimilar and 5 being very similar. Results for TPRL: 5 (9\%), 4 (43\%), 3 (33\%), and 2 (15\%); SCPN: 5 (9\%), 4 (36\%), 3 (29\%), 2.5 (1\%), 2 (23\%), and 1 (1\%); StyleAdv: 4 (22\%), 3 (33\%), 2 (34\%), and 1 (11\%). TPRL achieved the highest similarity ratings in categories 5, 4, and 3, indicating similarity to the original samples. SCPN ranked second, while StyleAdv received the lowest ratings. These GPT-3.5 findings align with human evaluation results.

\subsection{Measuring Diversity Via GPT-3.5}
To validate our observation regarding the diversity of the generated samples, we used GPT-3.5 to assess the diversity of generated samples for the three baselines. Using “from 1 to 5, how much is the generated sentence diverse from the original (1 being very non-diverse and 5 being very diverse)?” as a prompt. We randomly selected 250 sentences from SST-2. TPRL had 100 samples, SCPN had 77 samples, and StyleAdv had 73 samples. The scaling rates were as follows: TPRL: 5 (6\%), 4 (34\%), 3 (45\%), 2 (12\%), and 1 (3\%). SCPN: 5 (11\%), 4 (20\%), 3 (19\%), 2 (44\%), and 1 (3\%). StyleAdv: 4 (9\%), 3 (15\%), 2 (54\%), and 1 (20\%). These results confirm that TPRL generates more diverse samples, while cosine similarity fails to account for this diversity and considers it as a high similarity.

\section{Classifiers Error Analysis}\label{appendix:ceanaly}
To demonstrate dissimilarities in errors across employed classifiers, we utilized the following methodology: We inspected the intersection of misclassified samples for each dataset to examine whether or not a sample was present in all classifiers' misclassification sets, which we have termed the AND operation. Additionally, to inspect whether a sample was present in any of the classifiers' misclassification sets, we searched for unique samples, which we have designated as the OR operation. Our analysis indicates that the AND operation ranges from 9\% to 16\%, with an average of 10.57\%. Conversely, the OR operation ranges between 32\% and 55\%, averaging 41.49\%. Following fine-tuning with transferred samples from differing classifiers, we evaluated whether the improved performance was solely achieved through shared samples, which yielded a shared sample average of 30\%. Our analysis confirms that the policy shares both universal and model-specific features.


\section{Human Annotation Details
}\label{appendix:had}
We further conduct a human evaluation study of our attacks to examine to what extent are adversarial texts generated by TPRL truly imperceptible. We asked the annotators to follow the following instructions:\\

In this task, you will have two sentences, and you are required to say whether it has the same semantic meaning or not.\\

Semantic meaning in this context means that the two sentences have the same meaning, may be fully or partially. Let's look at the following example:\\

Ex.1.\\
S1: Ezekiel Ansah is wearing 3D glasses wout the lens,\\ 
S2: Wait Ezekiel ansah is wearing 3d movie glasses with the lenses knocked out.\\
Same Meaning: YES\\ 

Ex.2.\\
S1: Marriage equality law passed in Rhode Island,\\ 
S2: Congrats to Rhode Island becoming the 10th state to enact marriage equality.\\
 Same Meaning: YES\\

EX.3.\\
S1. Finally saw the Ciara body party video\\ 
S2. ciara s Body Party video is on point\\ 
Same Meaning: NO\\ 

Ex.4.\\ 
S1. Now lazy to watch Manchester united vs arsenal\\ 
S2. Early lead for Arsenal against Manchester United\\
Same Meaning: NO\\ 

In the first sentence, the two sentences are fully equivalent. In the second sentence, the two sentences are partially equivalent. In the third and fourth sentences, the two sentences are fully not equivalent. So, if the sentences are fully/partially equivalent, we can consider them as the same meaning.

\section{Hardware \& Software Dependencies}\label{appendix g}
For the paraphraser fine-tuning process, we utilized a cluster equipped with 4x V100 GPUs, each with 32GB of memory. To enhance the efficiency, we employed a zero-2 stage DeepSpeed framework \cite{rajbhandari2020zero}. These models were fine-tuned using the HuggingFace library \cite{wolf2019huggingface} and PyTorch \cite{paszke2017automatic}.
For RL fine-tuning, we utilized 2x V100 GPUs, each with 32GB of memory, and employed TRL (Transformer Reinforcement Learning) library\cite{vonwerra2022trl}.
\end{document}